\newif\ifarxiv
\arxivtrue

\ifarxiv
  \PassOptionsToPackage{table}{xcolor}
  \documentclass{fairmeta}
  \DeclareFontShape{T1}{optimistic}{b}{n}{<-> s * [0.88] assets/optimistic}{}
  \DeclareFontShape{T1}{optimistic}{bx}{n}{<-> ssub * optimistic/b/n}{}

  \newcommand{\teaserwidth}{0.92\textwidth}
\else
  \documentclass{article} 
  \usepackage{iclr2027_conference,times}
  \newcommand{\teaserwidth}{\textwidth}
\fi

\usepackage{amsmath,amsfonts,bm}

\def\eqref#1{equation~\ref{#1}}

\def\1{\bm{1}}

\DeclareMathAlphabet{\mathsfit}{\encodingdefault}{\sfdefault}{m}{sl}
\SetMathAlphabet{\mathsfit}{bold}{\encodingdefault}{\sfdefault}{bx}{n}

\usepackage{hyperref}
\usepackage{url}
\usepackage{booktabs}
\usepackage{amsmath}
\usepackage{graphicx}
\usepackage[table]{xcolor}
\usepackage{longtable}
\usepackage{listings}
\usepackage{multirow}
\newcommand{\colorsquare}[1]{%
  \raisebox{0.05ex}{\textcolor{#1}{\rule{1.1ex}{1.1ex}}}%
}
\definecolor{gapcolor}{HTML}{327DD8}
\definecolor{innovatorcolor}{HTML}{27B381}
\definecolor{writercolor}{HTML}{EC703E}

\definecolor{trainedhl}{HTML}{1F5FA8}
\definecolor{basehl}{HTML}{B04A2F}

\definecolor{ourbase}{HTML}{FBEBD3}
\definecolor{ourtrained}{HTML}{FFCF86}

\lstdefinestyle{judgeprompt}{
  basicstyle=\ttfamily,
  breaklines=true,
  breakatwhitespace=false,
  columns=fullflexible,
  keepspaces=true,
  showstringspaces=false,
  frame=single,
  framerule=0.2pt,
  xleftmargin=0.4em,
  xrightmargin=0.4em,
  aboveskip=0.35em,
  belowskip=0.55em
}

\newcommand{\sysname}{IdeaScientist}

\title{IdeaScientist: Orchestrating Agents for Grounded Scientific Ideation}

\newcommand{\paperabstract}{%
Despite rapid progress in automating scientific research, generating promising and well grounded research solutions remains a central challenge. We isolate research ideation as a standalone task and build our solution on the intuition that a challenge in one field can often be addressed by a mechanism that solved an analogous challenge in another. Accordingly, we introduce \sysname{}, which decomposes ideation into gap finding, innovation, and report writing, and trains each role with reinforcement learning. These roles identify limitations in related work, draw solution intuitions from analogous problem settings, and develop those intuitions into complete research proposals. To facilitate discovery of insights across domains, we construct the Svalbard Idea Vault, a corpus of 2.77M decomposed research ideas for retrieval, training, and temporally controlled evaluation. Our evaluation restricts access to literature available before a cutoff date and assesses how closely proposed directions align with those later explored in 15K papers authored by human researchers. On Qwen3.6-27B, IdeaScientist outperforms the strongest open-source autoresearch baseline by 14.0\%, driven mainly by gains in novelty. On this 27B open backbone, IdeaScientist even outperforms Claude Code SDK with Claude-4.8-Opus and Codex SDK with GPT-5.4, by up to 5.9\%. \ifarxiv\else Code and data will be released upon acceptance.\fi%
}

\ifarxiv
  \author[1,*]{Jiarui~Liu}
  \author[2]{Renjie~Tao}
  \author[2]{Yiwei~Liao}
  \author[3]{Chuanyang~Jin}
  \author[2]{Kai~Sun}
  \author[2]{Xiao~Yang}
  \author[2]{Xinyuan~Zhang}
  \author[2]{Xilun~Chen}
  \author[2]{Zhuangqun~Huang}
  \author[4]{Lechen~Zhang}
  \author[5]{Yongjin~Yang}
  \author[6]{Yinghui~He}
  \author[7,8]{Weihao~Xuan}
  \author[2]{Rakesh~Wanga}
  \author[2]{Anuj~Kumar}
  \author[1]{Mona~T.~Diab}
  \author[3]{Wen-tau~Yih}
  \author[2]{Xin~Luna~Dong}

  \affiliation[1]{Carnegie Mellon University}
  \affiliation[2]{Meta Reality Labs}
  \affiliation[3]{FAIR at Meta}
  \affiliation[4]{University of Illinois Urbana-Champaign}
  \affiliation[5]{University of Toronto}
  \affiliation[6]{Princeton University}
  \affiliation[7]{The University of Tokyo}
  \affiliation[8]{RIKEN AIP}

  \contribution[*]{Work done during an internship at Meta}

  \date{\today}
  \correspondence{Jiarui Liu at \email{jiaruil5@andrew.cmu.edu}}
  \metadata[Code]{\url{https://github.com/jiarui-liu/IdeaScientist}}
  \metadata[Dataset]{\url{https://huggingface.co/datasets/Jerry999/SvalbardIdeaVault}}

  \abstract{\paperabstract}
\else
  \author{Anonymous authors\\ Paper under double-blind review}
\fi

\ifarxiv\else
\fi
\begin{document}

\maketitle

\ifarxiv\else
\begin{abstract}
\paperabstract
\end{abstract}
\fi

\section{Introduction}
\label{sec:intro}
As coding agents take on a growing share of implementation work, \textit{Autonomous Research (AutoResearch)} has attracted increasing attention \citep{lu2024ai,yamada2025ai,gottweis2026accelerating,schmidgall2025agent}. Yet most of this progress sits downstream: existing systems couple solution proposal with implementation, experimentation, and writing, making it difficult to determine whether success comes from strong research ideas or strong downstream execution. Frontier labs report the same asymmetry: OpenAI recently reported its ``automated research intern'' focuses on well-defined tasks under human direction, whereas high-level planning remains a minimal fraction of agent output \citep{openai2026research}.

In this paper we focus on the problem of \textit{research ideation}, which we define as follows: given a problem definition (the task input and the desired output) together with a \textit{challenge} that current solutions fail to meet, produce a grounded research proposal that addresses the challenge. The challenge may concern any desideratum of the output or of the process producing it, such as inferior quality, high inference latency, or privacy risks. We ask two questions: (1) \textit{how can agents generate high-quality research proposals for open research problems?} (2) \textit{how can such proposals be evaluated reliably \emph{before} experimentation,} when many plausible solutions exist and none is uniquely correct?

Where, then, do research ideas come from? Our key intuition is rooted in a recurring view in {\em Cognitive Science} and {\em Psychonomic Science}: novelty emerges by connecting existing knowledge in new ways, and a solution developed for one problem transfers to a seemingly unrelated one when the two share the same underlying structure \citep{mednick1962associative,uzzi2013atypical,gentner1983structure,gick1983schema,loewenstein1999analogical}. We thus draw insights {\em not} from prior work in the same domain, but from work \textit{addressing a similar challenge in a different field}. For example, web search cuts latency by low-cost retrieval of a candidate set before applying a sophisticated ranker; the same two-stage structure answers the latency challenge in recommendation, where first-stage filtering precedes fine-grained scoring.

We introduce \sysname{}, an agent framework that operationalizes this process with three specialized sub-agents. A \emph{gap finder} examines closely related prior solutions in the target domain to identify the shared methodological limitation that leaves the challenge unaddressed, or even amplifies it. An \emph{innovator} searches for mechanisms that have addressed a similar challenge in other fields, then adapts them to the target setting. Finally, a \emph{report writer} turns the resulting intuitions into a complete research proposal \citep{baek2025researchagent,keya2026sci}. We train each sub-agent with asynchronous GRPO and role-specific, rubric-based rewards.

\begin{figure}[tbp]
\centering
\includegraphics[width=\teaserwidth]{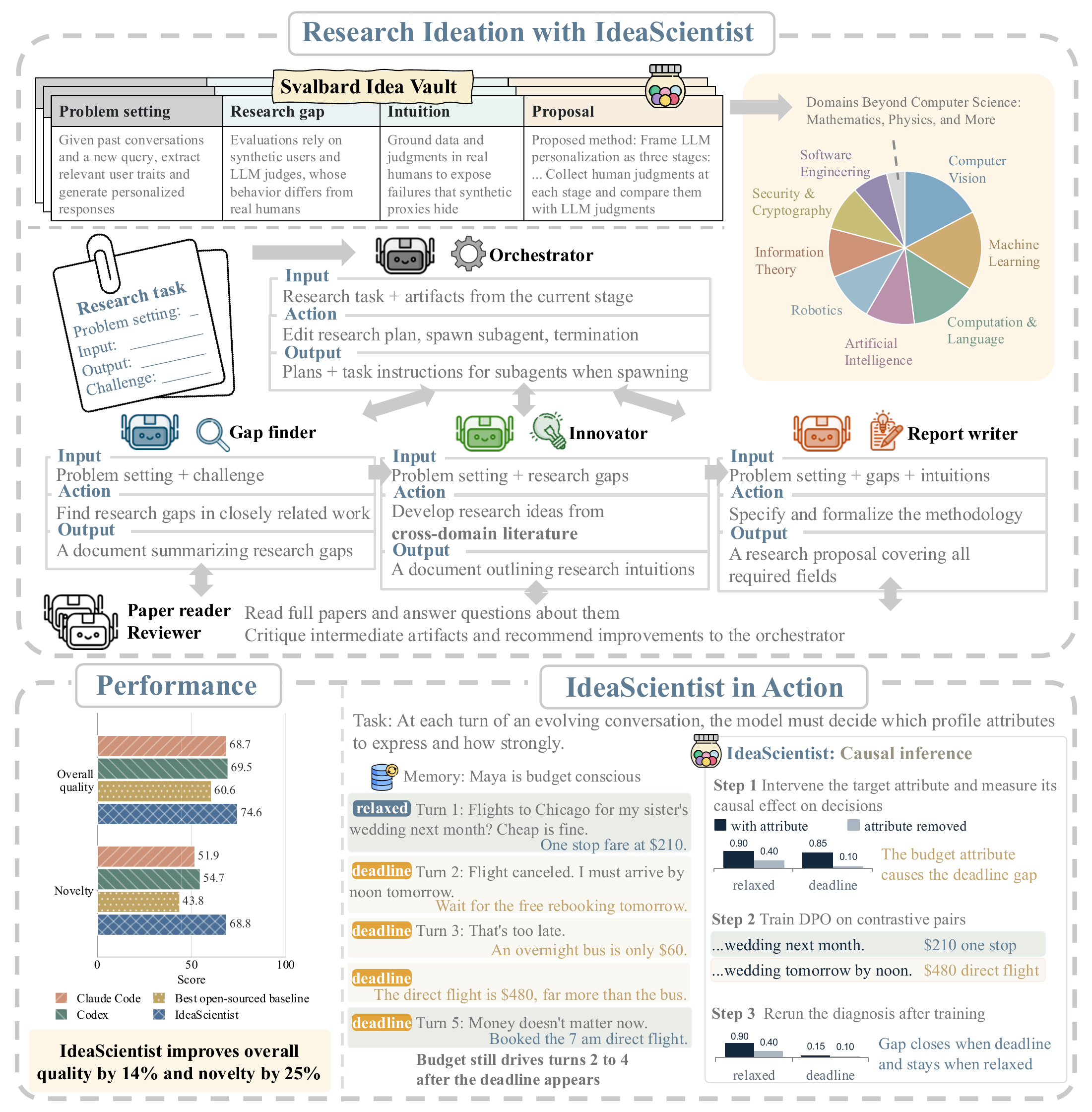}
\caption{Overview of \sysname{}. An orchestrator coordinates specialized agents---gap finder, innovator, and report writer---through shared intermediate artifacts, and the agents access the Svalbard Idea Vault for across-domain insights.}
\label{fig:harness}
\end{figure}

Both training and inference rely on a \textit{Svalbard Idea Vault}\footnote{This is named after the \textit{Svalbard Global Seed Vault}, as a metaphor for the repository of research ideas.}, an offline corpus built through \emph{idea decomposition}, which extracts 2.77 million (problem definition, challenge, research gap, intuition, proposal) tuples from research papers. We set the cutoff date to January 1, 2026, and treat the 14{,}977 methodology papers published after the cutoff as \textit{post-cutoff targets}. The idea vault serves three purposes. Pre-cutoff papers form the literature that the gap finder and the innovator search. A portion of the post-cutoff targets generates training data: the agent receives only the problem definition and challenge and produces a proposal, and comparing it against the target paper's actual solution yields the supervision signal for reinforcement learning. The remaining targets are held out for evaluation---because the agent cannot search literature published after the cutoff during evaluation, this design tests whether it can infer a research direction before that direction appeared.

In summary, this paper makes the following contributions\ifarxiv\else\footnote{We will release the data and code upon acceptance of the paper.}\fi: 
\begin{enumerate}
\item We separate out research ideation from AutoResearch, and formalize it as producing a grounded proposal from a problem definition and a challenge.
\item We introduce \sysname{}, an agent framework built on the intuition of searching for mechanisms that address similar challenges in different fields, and apply role-specific reinforcement learning to improve gap finding, innovation, and report writing.
\item We construct the Svalbard Idea Vault, an offline corpus of 2.77 million (problem definition, challenge, research gap, intuition, proposal) tuples produced through idea decomposition. The Idea Vault is used for literature search, training data generation, and evaluation.
\item We conduct extensive experiments showing the benefits of \sysname{}. On Qwen3.6-27B, it outperforms the strongest state-of-the-art autoresearch baseline by 14.0\% in overall score, which averages novelty and proposal quality. On the same open backbone, it also outperforms Claude Code SDK with Claude-4.8-Opus by 5.9\% and Codex SDK with GPT-5.4 by 5.1\%.
\end{enumerate}
We believe \sysname{} takes an important step toward fully automated research, and that it can be combined with automated research-topic identification and automated experimentation for stronger end-to-end auto-research.

\section{Related Work}
\label{sec:related}

\subsection{Autonomous Research Systems and Research Ideation}
\label{sec:related-systems}

End-to-end autoresearch systems combine ideation with implementation through agent workflows, experiment search, debate, or persistent memory \citep{lu2024ai,baek2025researchagent,ghafarollahi2025sciagents,ifargan2025autonomous,schmidgall2025agent,liu2026autoresearchclaw,song2026dr,tang2026ai,lu2026towards,yamada2025ai,lyu2026evoscientist,jin2026toward,wang2026idea,gottweis2026accelerating,su2025many,weng2026deepscientist,mitchener2025kosmos}. Their objectives typically reward experiment selection or downstream success, entangling proposal quality with coding and empirical outcomes \citep{falck2026training,yang2026scaling}. Recent work supervises ideation more directly through execution feedback, supervised fine-tuning followed by rewards for technical depth and semantic alignment, debate-derived rewards, or paper-specific reference-free rubrics \citep{zhang2026learning,gu2026mori,salimi2026debate,wang2026papergym}. However, these methods do not train literature search as an inspectable sequence of research operations.

Retrieval-based deepresearch systems structure evidence use through knowledge graphs, hybrid retrieval, citation paths, iterative search, or outline-guided synthesis \citep{wang2024scimon,wang2024scipip,li2025chain,hu2024nova,wang2024autosurvey,yan2025surveyforge,shao2024assisting}. Other methods rank research gaps or recombine ideas across scholarly trajectories \citep{shahid2025literature,xu2025can,li2026tree,radensky2024human,suzuki2026tca,choi2026read}. \sysname{} combines these threads in an inspectable ideation process, separately training agents to identify limitations in closely related work, seek inspiration from how other fields address similar challenges, and develop those insights into complete proposals.

\subsection{Evaluating Generated Research Ideas}
\label{sec:related-eval}

Pre-execution evaluation must substitute observable proxies for eventual scientific value, but different proxies answer different questions. Reference-free rubrics use LLM judges to assess whether an idea appears high quality across dimensions such as novelty and feasibility \citep{guo2025ideabench,ruan2026evaluating,chen2026ideation,schopf2026idea,wang2026proves}. Learned reviewers estimate acceptance-like quality \citep{zhu2025deepreview,idahl2025openreviewer,weng2025cycleresearcher,zeng2025reviewrl,nguyen2026benchmarking,deng2026cocoreviewbench,li2026llm,camelo2026closely,akella2025pre,wen2026predicting}. Factuality metrics test whether generated claims are directly supported by cited or retrieved scientific evidence \citep{min2023factscore,song2024veriscore,dhuliawala2024chain}. However, LLM judges can disagree, exhibit self-preference, or reward stylistic features \citep{zheng2023judging,panickssery2024llm,li2026can,ji2026style,norman2026reliability,sinhahajari2026limits, zhang2026re}. Expert studies also find that greater novelty may come at the cost of feasibility in actual execution \citep{si2025can,ho2026soundnessbench,si2026ideation,jiang2026hindsight}. Direct execution offers a more direct outcome-based signal, but it is costly and mixes ideation with implementation skill \citep{foster2026ai}. We validate and justify our evaluation metrics through cross-evaluation and manual annotation.

\subsection{Training LLM Agents for Non-Verifiable, Long-Horizon Tasks}
\label{sec:related-rl}

Ideation is difficult to train because unlike code, a proposal has no directly verifiable reward. Rubric-based reinforcement learning converts qualitative criteria into scalar supervision \citep{gunjal2026rubrics,huang2025reinforcement,zhou2025breaking,arora2025healthbench,mahmoud2026reward,liu2026openrubrics,xu2025direct,shao2025dr,goel2025training}, but a single proposal-level rubric provides weak credit assignment and invites reward hacking \citep{wang2026reproducing,yang2026rubric}. Outcome feedback can optimize long tool-use trajectories \citep{jin2025search,chen2025research,luo2025deepswe,wei2026swe}. In multi-agent settings, however, prior methods either specialize roles under a shared terminal objective or distill their trajectories into one policy \citep{motwani2024malt,zhao2026sirius,park2025maporl,li2026flow,li2025chainof}. This leaves unclear which reasoning stage improved and whether specialization survives deployment. To make long-horizon research ideation trainable, we use Group Relative Policy Optimization (GRPO) \citep{shao2024deepseekmath} to train each subagent role with a stage-matched human reference artifact and rubric, then recompose the trained roles within the full agentic loop.
\section{\sysname{}: A Framework for Autoresearch Ideation}
\label{sec:harness}

\subsection{Harness Design}
\label{subsec:harness-design}
\sysname{} decomposes research ideation into three specialized sub-agents coordinated by an orchestrator: a \textit{gap finder} examines closely related works to identify methodological limitations, an \textit{innovator} draws on work from different problem settings to develop transferable solution intuitions, and a \textit{report writer} consolidates the surviving intuition into a complete proposal. The report writer follows a fixed output schema that requires it to formalize the problem setting, state the central novelty, distinguish the proposed mechanism from the closest prior methods, and provide sufficient architectural and training detail to make the proposal actionable. It also specifies an evaluation plan, falsifiable predictions and their refutation conditions, and the proposal's limitations, while supporting claims about prior work with inline database identifiers. Table~\ref{tab:schema-reportwriter} provides the complete schema.

The orchestrator maintains the evolving research plan and determines which sub-agent to invoke next, allowing the system to revisit earlier stages when additional evidence or refinement is needed. The specialized agents execute the corresponding stages of research ideation through predefined interfaces.

Because full papers are often too long to include directly in every agent context, we separate planning and synthesis from full paper reading and reviewing. A dedicated second-tier \textit{paper reader} accesses full paper text in a separate context, performs targeted QA, and returns its responses to the requesting agent. A second-tier \textit{reviewer} evaluates quality of generated research intuitions and proposals.

Figure~\ref{fig:harness} shows the resulting flow from the problem and challenge to a grounded gap, a research intuition that may be adapted from another field, and a complete proposal. These intermediate artifacts let us check and train the gap finder, innovator, and report writer separately. The complete action space for each role is listed in Appendix~\ref{sec:appendix-action-space}.

\subsection{Training Design}
\label{sec:training}

We train the gap finder, innovator, and report writer separately before recomposing them in the shared harness, while preserving the same inputs, outputs, information access, and actions during training and deployment. Per-role training is possible because each agent has a clearly defined input and output, while human-authored papers provide valuable information from which corresponding reference artifacts can be reconstructed. This decomposition turns a sparse proposal-level outcome into stage-specific learning signals without prescribing a single tool-use trajectory. We optimize a separate LoRA adapter for each role using asynchronous Group Relative Policy Optimization (GRPO) \citep{shao2024deepseekmath}. We keep the orchestrator fixed because its deliberately narrow action space to maintain the research plan and schedule specialized roles has little room for improvement. Appendix~\ref{sec:appendix-setup} reports the training configuration details.

Rather than relying only on reference-free LLM judgments, we design a reference-grounded reward that evaluates both the standalone quality of each generated artifact and its agreement with the corresponding human reference. The resulting reward provides an outcome-level learning signal for the complete trajectory of decisions and tool calls that produces the artifact. We use three complementary reward components.
\begin{itemize}
\item Reference-free score $r_{\mathrm{free}}$: An LLM judge rates the specificity, methodological significance, feasibility, and evidential support of the generated artifact, measuring whether it is well formed and coherent independently of the reference.
\item Reference-grounded score $r_{\mathrm{ref}}$: A field-level comparison measures agreement with the reference on substantive research decisions such as the core limitations, proposed method, evaluation design, and related work.
\item Citation F1 $r_{\mathrm{cit}}$: Deterministic set F1 measures overlap between the papers cited by the generated artifact and those cited by the reference, capturing alignment in literature grounding and retrieval.
\end{itemize}
The rubric reward combines these three components subject to a validity gate:
\begin{equation*}
r_{\mathrm{rubric}} = g \cdot \left( \lambda_{\mathrm{free}} r_{\mathrm{free}} + \lambda_{\mathrm{ref}} r_{\mathrm{ref}} + \lambda_{\mathrm{cit}} r_{\mathrm{cit}} \right),
\end{equation*}
where all three weights are $1/3$ in the reported runs. The gate \(g \in \{0,1\}\) is determined by structural, task-compliance, citation-grounding, and tool-validity checks, together with a judged methodological criterion for the gap finder and innovator.

Finally, the training reward is $r=0.85r_{\mathrm{rubric}}+0.15r_{\mathrm{proc}}$, where $r_{\mathrm{proc}}$ is a rule-based term that penalizes protocol violations (e.g., disallowed tool calls, invalid formats, missing required fields). Appendix~\ref{sec:appendix-training} provides detailed rubric definitions and reward design choices.

\section{Dataset Construction}
\label{sec:data}

\subsection{Idea Vault construction}
We constructed the Svalbard Idea Vault as an offline literature database using the official arXiv API. The database covers all arXiv subject areas to allow drawing on relevant work from neighboring fields. The collection contains papers published from 1986 through June 18, 2026, with duplicates removed and full text collected whenever possible. For papers collected on or after January 1, 2026, we additionally follow their reference lists one hop and add cited papers that are missing from the database, retrieving full text whenever available. The resulting database contains 4.7 million papers, including 2.8 million (60\%) with full text.

For each paper, we apply \textit{idea decomposition} to reconstruct the problem definition and challenge as inputs and three chained reference artifacts as targets: the methodological gap, the proposed research intuition, and the complete research proposal. These artifacts correspond to the outputs of the gap finder, innovator, and report writer.

\subsection{Training data construction}
We construct training examples from machine learning papers published on or after January 1, 2026. Starting from an initial corpus of 104K candidate papers, we first retained papers whose LaTeX source is available for download and for which more than $70\%$ of cited works have usable full text in our database. We then restricted the corpus to methodological papers, such as contributing a new method, model, algorithm, architecture, training procedure, or technical framework, excluding surveys, position papers, datasets and benchmarks, and purely empirical engineering reports.

These filters yielded $14{,}977$ target papers. We randomly partitioned them into $14{,}500$ training papers, $200$ validation papers, and $277$ held-out test papers, and used the same partition for all three agents. Qwen3.6-27B converts each human-authored target paper into the required proposal schema, preserving the research problem, proposed method, model design, related work, and limitations, while removing all measured experimental outcomes. This results-masked representation records directions that human researchers have pursued without exposing their experimental success.

We generate training data using the extracted gaps, intuitions, and research proposal, which provides the skeleton about the research direction; supporting claims and citations must be grounded in papers read through the environment. We validate every artifact carefully before adding to the training set. Appendix~\ref{sec:appendix-label-generation} reports details of the training reference construction.

\section{Experiments}
\label{sec:experiments}

\subsection{Experiment Setup}
\label{sec:setup}

We evaluate \sysname{}-Base and \sysname{}-Trained on Qwen3.6-27B. In the trained variant, the gap finder, innovator, and report writer use their trained checkpoints. Every run uses a fixed generation token budget per turn and a fixed turn limit for each role.

We compare against three families of methods: (1) general-purpose research agents, including Claude Code with Claude Opus 4.8 and Codex with GPT 5.4; (2) open-source autoresearch systems including end-to-end systems such as AutoResearchClaw~\citep{liu2026autoresearchclaw}, Dr.~Claw~\citep{song2026dr}, and DeepScientist~\citep{weng2026deepscientist}, and research ideation systems including ResearchAgent~\citep{baek2025researchagent}, CycleResearcher paired with its DeepReviewer critic~\citep{weng2025cycleresearcher,zhu2025deepreview}, SciAgents~\citep{ghafarollahi2025sciagents}, and AI-Researcher~\citep{tang2026ai}, with different backbones including Qwen3.6-27B, Gemma-4-31B, and GPT-OSS-120B; (3) \sysname{}-Base (our harness) on the same backbones, and \sysname{}-Trained.

For a fair comparison, all systems receive the same problem definition and output schema and use the same tools to access our offline paper database for literature retrieval, while the target paper remains hidden and internet access is disabled. External AutoResearch systems and general-purpose agents retain their native agent loops. For end-to-end research systems, we instruct the agent to stop once a machine learning research proposal is complete rather than continuing to implementation or experimentation. Appendix~\ref{sec:appendix-setup} reports the full configuration.

\paragraph{Evaluation Metrics.}
\label{sec:eval}
\label{sec:eval-endtoend}
We evaluate the generated research proposal for each held-out target paper using two complementary groups of metrics.
First, \textit{Novelty} measures how difficult it is to derive the proposal's central mechanism from the comparison literature in the Idea Vault. We compare with three scopes of literature: against all literature (All), against literature before the cutoff date (Cutoff), and against literature in the same problem domain.
Second, \textit{Proposal quality} measures the research proposal on six criteria commonly used in conference peer review: \textit{relevance, clarity, specificity, actionability, soundness,} and \textit{impact}. Appendix~\ref{sec:appendix-evaluation-prompts} gives the judgment prompts.

\subsection{Q1. Can Our Agents Generate High-Quality Research Proposals?}
\label{sec:results}

\begin{table}[t]
\centering
\small

\caption{Proposal-level evaluation on the held-out 277-paper test split. We measure novelty against all literature (All), pre-cutoff literature (Cutoff), and same-domain literature (In-domain), with their mean (Avg). We measure proposal quality along six dimensions: with relevance (Relev.), clarity (Clar.), specificity (Spec.), actionability (Act.), soundness (Sound.), and impact (Impact), with their mean (Avg). The overall score is the mean of the novelty Avg and the proposal-quality Avg. All metrics are scored by a Qwen3.6-27B judge. Within each group, the best result is bolded and the second-best is underlined. \sysname{} outperforms state-of-the-art systems by up to 14\%, especially on novelty.}

\label{tab:endtoend}

\resizebox{\textwidth}{!}{%
\begin{tabular}{@{}lcccccccccccc@{}}
\toprule

\multirow{2}{*}{System}
& \multicolumn{4}{c}{Novelty}
& \multicolumn{7}{c}{Proposal quality}
& \multirow{2}{*}{\textit{Overall}}
\\

\cmidrule(lr){2-5}
\cmidrule(lr){6-12}

&
All
& Cutoff
& In-domain
& \textit{Avg}
&
Relev.
& Clar.
& Spec.
& Act.
& Sound.
& Impact
& \textit{Avg}
&
\\

\midrule

\multicolumn{13}{@{}l}{\textit{Claude-4.8-Opus}}\\

Claude Code SDK
& 68.3 & 70.4 & 17.0 & \textit{51.9}
& 98.2 & \textbf{89.9} & 89.2 & \textbf{84.7} & \textbf{73.7} & 77.4 & \textit{85.5}
& \textit{68.7}
\\

\rowcolor{ourbase}
\textbf{\sysname{}-Base}
& \textbf{77.6} & \textbf{77.0} & \textbf{54.9} & \textbf{\textit{69.8}}
& \textbf{99.5} & 89.6 & \textbf{90.8} & 83.5 & 72.2 & \textbf{78.3} & \textbf{\textit{85.7}}
& \textbf{\textit{77.7}}
\\

\midrule

\multicolumn{13}{@{}l}{\textit{GPT-5.4}}\\

Codex SDK
& \textbf{72.4} & \textbf{71.8} & 19.9 & \textit{54.7}
& 97.1 & 87.4 & \textbf{88.5} & \textbf{83.4} & \textbf{73.6} & \textbf{76.1} & \textbf{\textit{84.4}}
& \textit{69.5}
\\

\rowcolor{ourbase}
\textbf{\sysname{}-Base}
& 67.8 & 67.5 & \textbf{47.7} & \textbf{\textit{61.0}}
& \textbf{98.2} & \textbf{91.3} & 82.1 & 78.4 & 71.6 & 75.0 & \textit{82.8}
& \textbf{\textit{71.9}}
\\

\midrule

\multicolumn{13}{@{}l}{\textit{Open-source systems on Gemma-4-31B}}\\

AutoResearchClaw
& 52.6 & 51.9 & 21.7 & \textit{42.1}
& \textbf{98.8} & 85.6 & 67.7 & 64.6 & 53.5 & \underline{70.3} & \textit{73.4}
& \textit{57.7}
\\

Dr.\ Claw
& 50.3 & 50.8 & 13.7 & \textit{38.3}
& 94.4 & 85.8 & 69.9 & 69.6 & \underline{59.3} & 69.8 & \textit{74.8}
& \textit{56.5}
\\

ResearchAgent
& 52.1 & \underline{53.5} & 18.8 & \textit{41.5}
& 91.6 & 84.4 & 68.0 & 66.6 & 57.9 & 69.7 & \textit{73.0}
& \textit{57.3}
\\

CycleResearcher
& 46.0 & 45.2 & 10.8 & \textit{34.0}
& \underline{97.0} & \underline{86.2} & 70.3 & 70.0 & 57.9 & 68.7 & \textit{75.0}
& \textit{54.5}
\\

SciAgents
& 52.8 & 53.3 & \underline{54.9} & \underline{\textit{53.7}}
& 51.9 & 70.6 & 53.7 & 39.5 & 34.8 & 66.9 & \textit{52.9}
& \textit{53.3}
\\

AI-Researcher
& \underline{54.1} & 52.0 & 18.4 & \textit{41.5}
& 96.0 & 85.1 & \underline{72.7} & \underline{70.6} & 57.8 & 69.5 & \underline{\textit{75.3}}
& \underline{\textit{58.4}}
\\

DeepScientist
& 42.3 & 41.8 & 15.9 & \textit{33.3}
& 89.8 & 76.0 & 55.6 & 55.4 & 51.0 & 63.0 & \textit{65.1}
& \textit{49.2}
\\

\rowcolor{ourbase}
\textbf{\sysname{}-Base}
& \textbf{61.4} & \textbf{62.5} & \textbf{64.3} & \textbf{\textit{62.7}}
& 94.9 & \textbf{89.1} & \textbf{74.5} & \textbf{73.3} & \textbf{60.3} & \textbf{70.6} & \textbf{\textit{77.1}}
& \textbf{\textit{69.9}}
\\

\midrule

\multicolumn{13}{@{}l}{\textit{Open-source systems on GPT-OSS-120B}}\\

AutoResearchClaw
& 53.0 & 53.2 & 27.1 & \textit{44.4}
& \underline{96.6} & 84.4 & 75.6 & 67.6 & 50.4 & 67.4 & \textit{73.7}
& \textit{59.1}
\\

Dr.\ Claw
& 52.8 & 53.0 & 22.0 & \textit{42.6}
& 94.2 & \underline{85.3} & \underline{76.5} & \underline{71.8} & \textbf{56.2} & \underline{69.3} & \underline{\textit{75.6}}
& \textit{59.1}
\\

ResearchAgent
& 45.8 & 45.7 & 16.6 & \textit{36.0}
& 95.6 & 84.9 & 71.8 & 68.0 & 53.5 & 68.0 & \textit{73.6}
& \textit{54.8}
\\

CycleResearcher
& 48.4 & 47.0 & 18.8 & \textit{38.1}
& 90.6 & 80.7 & 70.2 & 64.5 & 51.8 & 66.5 & \textit{70.7}
& \textit{54.4}
\\

SciAgents
& 46.2 & 45.3 & \textbf{35.7} & \textit{42.4}
& 81.3 & 70.2 & 64.2 & 43.1 & 34.7 & 60.8 & \textit{59.1}
& \textit{50.7}
\\

AI-Researcher
& \underline{53.8} & \underline{54.7} & 28.5 & \underline{\textit{45.7}}
& 92.2 & 82.7 & \textbf{77.8} & 69.5 & \underline{55.2} & 68.2 & \textit{74.3}
& \underline{\textit{60.0}}
\\

DeepScientist
& 46.0 & 44.3 & 18.4 & \textit{36.2}
& 93.4 & 84.8 & 72.1 & 67.1 & 52.2 & 67.8 & \textit{72.9}
& \textit{54.6}
\\

\rowcolor{ourbase}
\textbf{\sysname{}-Base}
& \textbf{59.2} & \textbf{57.5} & \underline{31.3} & \textbf{\textit{49.3}}
& \textbf{98.2} & \textbf{88.1} & 76.4 & \textbf{73.5} & \textbf{56.2} & \textbf{71.1} & \textbf{\textit{77.3}}
& \textbf{\textit{63.3}}
\\

\midrule

\multicolumn{13}{@{}l}{\textit{Open-source systems on Qwen3.6-27B}}\\

AutoResearchClaw
& 52.2 & 50.9 & 18.4 & \textit{40.5}
& \textbf{99.4} & 86.6 & 69.6 & 67.2 & 54.2 & 70.7 & \textit{74.6}
& \textit{57.6}
\\

Dr.\ Claw
& 47.5 & 47.5 & 18.4 & \textit{37.8}
& 96.5 & 85.4 & 68.3 & 66.7 & 56.9 & 69.8 & \textit{73.9}
& \textit{55.9}
\\

ResearchAgent
& 46.1 & 44.5 & 11.2 & \textit{33.9}
& 97.5 & 86.2 & 70.8 & 68.7 & 56.5 & 70.0 & \textit{75.0}
& \textit{54.4}
\\

CycleResearcher
& 49.9 & 48.8 & 10.5 & \textit{36.4}
& 96.4 & 86.3 & 72.3 & 69.9 & 59.1 & 70.8 & \textit{75.8}
& \textit{56.1}
\\

SciAgents
& 44.7 & 44.3 & 9.0 & \textit{32.7}
& 96.4 & 85.7 & 68.7 & 66.8 & 55.6 & 69.9 & \textit{73.9}
& \textit{53.3}
\\

AI-Researcher
& 54.6 & 56.7 & 19.5 & \textit{43.6}
& 94.9 & 85.4 & 73.4 & 69.7 & 57.7 & 70.5 & \textit{75.3}
& \textit{59.4}
\\

DeepScientist
& 57.6 & 58.5 & 15.2 & \textit{43.8}
& 94.0 & 85.8 & 77.8 & 74.9 & 61.9 & 70.6 & \textit{77.5}
& \textit{60.6}
\\

\rowcolor{ourbase}
\textbf{\sysname{}-Base}
& \underline{66.0} & \underline{67.1} & \underline{67.5} & \underline{\textit{66.9}}
& 96.2 & \underline{89.9} & \textbf{80.6} & \underline{77.3} & \underline{62.8} & \underline{73.4} & \underline{\textit{80.0}}
& \underline{\textit{73.5}}
\\

\rowcolor{ourtrained}
\textbf{\sysname{}-Trained}
& \textbf{67.3} & \textbf{68.3} & \textbf{70.9} & \textbf{\textit{68.8}}
& \underline{98.2} & \textbf{90.3} & \underline{77.9} & \textbf{77.8} & \textbf{64.5} & \textbf{73.5} & \textbf{\textit{80.4}}
& \textbf{\textit{74.6}}
\\

\bottomrule
\end{tabular}}
\end{table}

Table~\ref{tab:endtoend} evaluates the ideation systems using the metrics defined in Section~\ref{sec:eval-endtoend}. Even without subagent training, \sysname{}-Base is competitive on every backbone. Against general-purpose agents running in their native SDK harnesses, \sysname{}-Base leads Claude Code SDK on Claude-4.8-Opus by 9.0\% on the overall score (77.7\% vs.\ 68.7\%) and leads Codex SDK on GPT-5.4 by 2.4\% (71.9\% vs.\ 69.5\%). Against state-of-the-art autoresearch systems, \sysname{}-Base ranks first on all three open backbones. Its margin over the strongest matched-backbone baseline ranges from 3.3\% on GPT-OSS-120B to 12.9\% on Qwen3.6-27B.

Our reward design enables training to improve all three producer roles. In isolated held-out evaluation, the final checkpoints raise the reward from 56.6\% to 66.7\% (+10.1\%) for the gap finder, from 58.8\% to 69.5\% (+10.7\%) for the innovator, and from 78.6\% to 89.1\% (+10.5\%) for the report writer (Appendix Figure~\ref{fig:reward_curves} and Table~\ref{tab:persubagent}). Recomposing the three trained roles carries these gains over to proposal generation. \sysname{}-Trained improves over \sysname{}-Base by 1.9\% on average novelty and 0.4\% on average proposal quality. It reaches an overall score of 74.6\%, which makes it the strongest open system and puts it 14.0\% above the best baseline on Qwen3.6-27B.

\begin{figure}[t]
\centering
\begin{minipage}[b]{0.4837\textwidth}
    \centering
    \includegraphics[width=\linewidth]{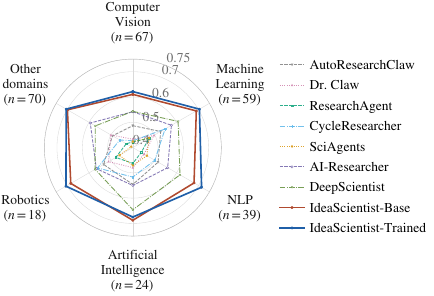}

    \footnotesize\textbf{(a)} Novelty across research domains
\end{minipage}\hfill
\begin{minipage}[b]{0.5013\textwidth}
    \centering
    \includegraphics[width=\linewidth]{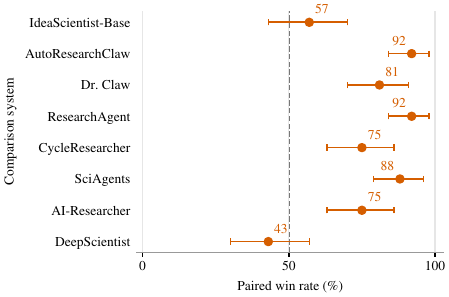}

    \footnotesize\textbf{(b)} \sysname{}-Trained vs. each system
\end{minipage}
\caption{Proposal-level results. \textbf{(a)} Proposal novelty across research domains for different systems on Qwen3.6-27B; the number of test papers per domain is given in parentheses. \textbf{(b)} Human paired win rate of \sysname{}-Trained against each system on matched papers. Quality averages all seven dimensions, and error bars show bootstrap $95\%$ confidence intervals.}
\label{fig:results-overview}
\end{figure}

Across dimensions, \sysname{}'s advantage comes mainly from novelty. On average proposal quality, \sysname{}-Base is on par with or modestly above the strongest baseline, ranging from $1.8\%$ on Gemma-4-31B to $+2.5\%$ on Qwen3.6-27B. On average novelty, its gains are much larger, up to $+23.1\%$ on Qwen3.6-27B. On that backbone, the gain holds against all literature ($+8.4\%$), against pre-cutoff literature ($+8.6\%$), and most strongly against same-domain literature ($+48.0\%$). Being novel relative to a proposal's own research area requires thinking outside that area's established approaches; this is arguably the hardest form of novelty, and it is where our method stands strongest.
Figure~\ref{fig:results-overview}(a) shows that this advantage spans all six domain groups, with the estimated \sysname{}-Trained profile exceeding Base in five. 

\paragraph{Human-annotation-based win-rate.} We ran a blind pairwise human evaluation on 50 randomly selected papers from the held-out test split. For each paper, two annotators compared the proposal from \sysname{}-Trained with proposals from other systems on the same backbone, including \sysname{}-Base. System names were hidden. Figure~\ref{fig:results-overview}(b) reports the win rates: \sysname{}-Trained beats every other state-of-the-art system by a statistically significant margin, with DeepScientist the only exception. \sysname{}-Trained also wins more often than \sysname{}-Base, although the difference is not statistically significant.

\paragraph{Case Studies.} 
We manually examined differences between proposals generated by different systems, with representative examples presented in Appendix~\ref{sec:appendix-case-studies}. \sysname{}-Trained generally makes a tighter connection between the stated failure and the mechanism used to address it. Its key decision rules follow from the structure of the problem, making the underlying assumptions and expected effects easier to test. In comparison, baseline methods introduce plausible control rules with weaker justification for why those rules address the failure. 
\subsection{Q2: How Much Does Each Element Contribute?}
\label{sec:ablations}

\begin{figure}[t]
\centering
\begin{minipage}[b]{0.2995\textwidth}
    \centering
    \includegraphics[width=\linewidth]{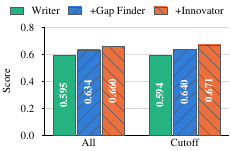}
\end{minipage}\hfill
\begin{minipage}[b]{0.2995\textwidth}
    \centering
    \includegraphics[width=\linewidth]{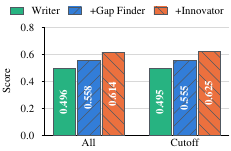}
\end{minipage}\hfill
\begin{minipage}[b]{0.3710\textwidth}
    \centering
    \includegraphics[width=\linewidth]{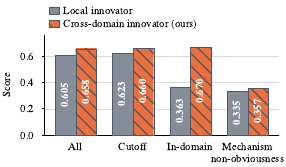}
\end{minipage}

\begin{minipage}[t]{0.2995\textwidth}
    \centering
    \footnotesize\textbf{(a)} Sub-agent ablation:\\[-0.2ex]
    Qwen3.6-27B
\end{minipage}\hfill
\begin{minipage}[t]{0.2995\textwidth}
    \centering
    \footnotesize\textbf{(b)} Sub-agent ablation:\\[-0.2ex]
    Gemma-4-31B
\end{minipage}\hfill
\begin{minipage}[t]{0.3710\textwidth}
    \centering
    \footnotesize\textbf{(c)} Innovator retrieval ablation:\\[-0.2ex]
    Qwen3.6-27B
\end{minipage}

\caption{Ablation study. \textbf{(a--b)} Sub-agent ablations: adding the Gap Finder and Innovator on top of the Report Writer improves All and Cutoff novelty by up to 7.7\% on Qwen3.6-27B and up to 13.0\% on Gemma-4-31B. \textbf{(c)} Innovator retrieval ablation: cross-domain retrieval improves novelty over same-domain retrieval, most strongly on In-domain novelty (+30.7\%).}
\label{fig:ablations}
\end{figure}

\paragraph{Sub-agent contributions.}
\label{sec:stage-ablation}
We first measure how much each sub-agent contributes to novelty. Figure~\ref{fig:ablations}(a--b) compares three configurations: the Report Writer alone (the minimal setup that can write a proposal), the Report Writer with the Gap Finder, and the full pipeline, which adds the Innovator. Each added sub-agent raises both All and Cutoff novelty scores, and the weaker backbone (Gemma-4-31B) benefits more from the pipeline. Identifying research gaps first and then innovating on them helps the system reach ideas that are hard to derive from the problem statement alone.

\paragraph{Innovator retrieval scope.}
\label{sec:retrieval-regime}
We next vary the Innovator's retrieval scope, comparing same-domain retrieval with cross-domain retrieval. As in Table~\ref{tab:endtoend}, we measure novelty against three scopes of literature. We add a fourth, stricter measure, \emph{Mechanism non-obviousness}: we prompt Qwen3.6-27B to independently derive candidate solutions from close prior work and score the proposal's novelty against these derived solutions. As Figure~\ref{fig:ablations}(c) shows, cross-domain retrieval improves every novelty measure: All (+5.3\%), Cutoff (+3.7\%), and most strongly In-domain (+30.7\%, from 36.3\% to 67.0\%). Mechanism non-obviousness is the hardest measure for both variants, but cross-domain retrieval still raises it from 33.5\% to 35.7\% (+2.2\%).

\subsection{Q3: Can We Evaluate Research Proposal Quality Reliably?}
\label{sec:judge-reliability}

\begin{figure}[t]
\centering
\begin{minipage}[b]{0.4208\textwidth}
    \centering
    \includegraphics[width=\linewidth]{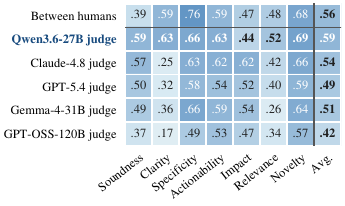}

    \footnotesize\textbf{(a)} Pearson correlation
\end{minipage}\hfill
\begin{minipage}[b]{0.2746\textwidth}
    \centering
    \includegraphics[width=\linewidth]{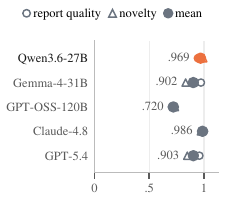}

    \footnotesize\textbf{(b)} Consensus agreement
\end{minipage}\hfill
\begin{minipage}[b]{0.2746\textwidth}
    \centering
    \includegraphics[width=\linewidth]{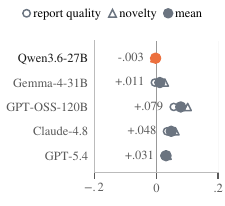}

    \footnotesize\textbf{(c)} Self-preference bias
\end{minipage}
\caption{Judge reliability supports our use of the Qwen3.6-27B judge. \textbf{(a)} Pearson correlation between the two human annotators, or between each judge and their averaged ratings; the last column averages across the seven dimensions. \textbf{(b)} System-level ranking agreement with the leave-one-out consensus of the other four judges. \textbf{(c)} Excess self-preference bias measures how much more a judge rates its own model family than the other judges do.}
\label{fig:judge-agreement}
\end{figure}

To check whether the LLM-judged quality scores are trustworthy, we compare them with scores from two human experts on 50 proposals generated by 9 systems. The two annotators scored the same quality dimensions as the LLM judges. Figure~\ref{fig:judge-agreement}(a) reports two Pearson correlations: between the two annotators on individual proposals, and between each LLM judge and the averaged human scores. We make two observations. First, scoring research proposals is inherently subjective, and even the two experts disagree to a notable degree ($r=0.56$), much as reviewers do in conference peer review. Second, each LLM judge correlates with the human average at a level comparable to the agreement between the two annotators ($r=0.51$--$0.59$ for top-3 judges). LLM judges are therefore a reasonable proxy for expert assessment.

Figure~\ref{fig:judge-agreement}(b--c) further measures two properties of each judge: its agreement with the cross-judge consensus, and its self-preference bias toward generation systems from its own model family. Qwen3.6-27B agrees strongly with the consensus, second only to the frontier Claude-4.8 judge, and the difference between the two is not statistically significant. Its self-preference bias is negligible and lower than Claude-4.8's, justifying why we chose it as the main judge. Appendix~\ref{sec:appendix-judge} reports report-level agreement and the full cross-judge breakdown.

\section{Conclusion}
\label{sec:conclusion}

\sysname{} isolates research ideation as a capability that can be trained and evaluated independently of downstream implementation. Our system identifies limitations in closely related work, searches other problem settings for mechanisms that may address the same challenge, and develops them into grounded research proposals. Using role-specific training with temporally held-out human papers as reference directions, \sysname{} improves each stage of ideation and becomes the strongest open system in our proposal-level evaluation. Our results suggest that structured agent environments can make scientific ideation amenable to controlled learning and evaluation, rather than treating it as an opaque byproduct of end-to-end autoresearch.

\ifarxiv\else
\subsection*{AI use statement}

We used generative AI tools only to polish phrasing and correct grammatical errors. We reviewed all AI-assisted edits and take responsibility for the final content of this work.
\fi

\bibliography{iclr2027_conference}
\bibliographystyle{iclr2027_conference}

\appendix
\section{Framework Details}

\subsection{Action Space and Retrieval}
\label{sec:appendix-action-space}
Each agent acts in turns. On a turn it may perform one or more permitted operations in parallel or hand control back. For the orchestrator, handing back control ends the run; for a sub-agent, it returns the result to the role that invoked it. Table~\ref{tab:agent-tools} summarizes these permissions in terms of what each role can access, modify, and return.

The orchestrator performs no literature retrieval or paper reading. It can spawn any producer or reviewer sub-agent, view any file a sub-agent wrote, and modify only the research plan. If it attempts to stop before the final proposal exists, it receives up to eight continuation prompts; the run is also bounded by a maximum turn count. Each producer and reviewer sub-agent can retrieve and request readings of papers and can write its working file. A producer that attempts to return before writing at least 20 non-whitespace characters to its required artifact receives up to three completion prompts. The reviewer has no analogous artifact gate. Each first-tier sub-agent may request at most 20 paper readings per task and is bounded by a role-specific turn limit. For the reported runs, the paper reader receives one paper's full text in a separate single-turn context and returns a digest, so full text never enters a first-tier agent's context.

\paragraph{Retrieval regimes.} Every retrieving role uses the same keyword-search interface over the offline corpus, but the roles are instructed to use it in the two different regimes described in \S\ref{subsec:harness-design}. The regime is therefore determined by how a role formulates queries rather than by a separate tool. The gap finder is given one challenge axis and instructed to issue four to eight queries phrased in the exact terminology of that axis and of the target problem definition. It re-queries with rephrasings while the returned pool is thin so that the same-challenge, same-problem-definition prior art is retrieved as completely as possible. Its gap schema then requires the near-miss papers and the reason each falls short (Table~\ref{tab:schema-gapfinder}), which is only satisfiable from that regime. The innovator is instructed to issue six to ten queries hunting for a mechanism whose logic could transfer to the assigned gap, preferring queries that leave the problem's own sub-field and span at least three distinct sub-fields. An in-domain source is admissible only when the transfer is non-trivial rather than a rediscovery of the closest prior work, and the resulting candidate must name the source paper, the mechanism, and the condition the transfer relies on. The reviewer searches in the gap finder's regime, since its task is to falsify the novelty claim against same-problem prior art. The report writer searches in whichever regime the claim it is grounding requires, because it must cite both the related work the proposal displaces and the source of the mechanism it proposes.

\begin{table}[t]
\centering
\caption{Role-specific information access, workspace permissions, and handoffs. Only the paper reader sees full paper text; all other roles use bibliographic metadata and task-focused reader digests.}
\label{tab:agent-tools}
\begin{tabular}{p{0.18\textwidth}p{0.34\textwidth}p{0.38\textwidth}}
\toprule
Role & Information access & Workspace and handoff \\
\midrule
Orchestrator & Reads the research plan and all intermediate artifacts; cannot search or read papers. & Updates only the research plan, invokes a producer or reviewer, and ends the run once the final proposal exists. \\
Gap finder, innovator, and report writer & Search the offline corpus through bibliographic metadata and request task-focused paper digests. & Each edits only its own stage artifact and returns the completed artifact to the orchestrator. \\
Reviewer & Searches bibliographic metadata and requests task-focused paper digests. & Writes review notes and returns its verdict to the orchestrator. \\
Paper reader & Receives one requested paper's full text in an isolated, single-turn context. & Answers the requesting role's questions and returns a focused digest. \\
\bottomrule
\end{tabular}
\end{table}

\subsection{Agent Output Schemas}
\label{sec:appendix-schemas}
Each producing role must write its artifact in a fixed schema, which is identical at training and deployment and is the surface the reward and the deterministic checks operate on. Tables~\ref{tab:schema-gapfinder} through \ref{tab:schema-reader} list the entries of each schema and the instruction attached to each entry. The gap finder appends one \texttt{Axis} section per assigned axis, containing a mechanism-level summary of what has been done and a list of at most two gaps, each following the gap schema. The innovator appends one candidate block per call, the reviewer appends one verdict block per reviewed candidate, and the report writer emits a single JSON object in the results-masked schema with every cited work referenced inline by its database id. The orchestrator's plan file is a free-form planning document and is not schema-constrained.

\begin{table}[t]
\centering
\caption{Gap finder output schema (\texttt{gaps.md}). Each assigned axis produces one section with a summary of prior work and at most two gaps. Every gap fills the entries below.}
\label{tab:schema-gapfinder}
\begin{tabular}{p{0.26\linewidth}p{0.66\linewidth}}
\toprule
Entry & Definition and instruction \\
\midrule
Gap & One sentence in contrast form: prior methods do X through Y, but Y leaves method-level issue Z unresolved, naming the specific task or setting and the method assumption being challenged. \\
Near-miss methods and coverage & The closest prior methods by paper id and how each comes close, the shared method-level deficiency across the papers read, and the ids or search trail checked, with at least two ids unless only one real near miss exists. \\
Methodological deficiency & The one precise methodological reason the near-miss work misses the gap, tagged by kind such as missing mechanism, algorithmic limitation, wrong inductive bias, credit-assignment flaw, training-objective mismatch, representation limitation, or scalability limit, grounded in a specific finding. \\
Why it matters & Who is blocked and what new understanding or capability closing the gap would unlock. \\
Evidence route & The observable evidence or comparison that would show a future method closes the gap, without proposing the solution. \\
\bottomrule
\end{tabular}
\end{table}

\begin{table}[t]
\centering
\caption{Innovator output schema (\texttt{candidates.md}). Each call appends one candidate block with the entries below.}
\label{tab:schema-innovator}
\begin{tabular}{p{0.26\linewidth}p{0.66\linewidth}}
\toprule
Entry & Definition and instruction \\
\midrule
Intuition & One or two sentences stating the single idea in plain language. \\
Gap it attacks & The specific gap from the gap analysis and the concrete failure of current approaches it addresses, and why closing it matters. \\
If improving prior candidate & Optional: which earlier candidate is being improved, the weakness addressed, and what changed. \\
Source inspiration & The source paper or papers by id and the mechanism or principle borrowed, why it works there, and the novelty type, one of new topic, recombination of distant concepts, or reinforcing a weak connection. \\
How it maps to this problem & A plain-language translation of the mechanism to this problem, without pseudocode, equations, architecture, or hyperparameters. \\
Why it could work & The reason the transfer is plausible given the source literature, and the key assumption that must hold. \\
Novelty vs. closest in-domain work & The closest in-domain paper by id and the specific difference from it. \\
Main risk & The single most likely reason the intuition is wrong or the transfer fails. \\
\bottomrule
\end{tabular}
\end{table}

\begin{table}[t]
\centering
\caption{Report writer output schema. The report is stored as a single JSON object in the results-masked schema; nested entries are summarized here. Every cited work is referenced inline by its database id, and no measured results are included.}
\label{tab:schema-reportwriter}
\begin{tabular}{p{0.26\linewidth}p{0.66\linewidth}}
\toprule
Entry & Definition and instruction \\
\midrule
Title & The proposal title. \\
Topic relevance & The sub-areas and primary focus the proposal addresses. \\
One-sentence thesis & A single sentence stating the contribution, under 30 words and without metric numbers. \\
Core problem & What fails and why, why it matters as motivation only, and a concrete failing example. \\
Key novelty & A short name for the contribution and a brief intuition-level explanation. \\
Problem definition & The formal setting, the inputs the system receives, and the outputs it produces. \\
Key terms & Plain-language definitions of the terms used. \\
Related work & Per sub-area, a mechanism-level summary of prior work, the open gap it leaves, and its citations. \\
Method & The high-level flow, the modules and their roles, novel architectural elements, key design choices, and any theoretical grounding. \\
Model details & The base model and the training approach, objectives, adaptation, and key hyperparameters the method needs. \\
Comparison to state of the art & The related methods and the specific differences from each. \\
Proposed evaluation & The evaluation setting, benchmarks, baselines, metrics, and ablations the method would use, stated as a plan without measured results. \\
Falsifiable predictions & Empirical claims the method implies, each paired with the observation that would falsify it. \\
Limitations & Three to five concise limitations, acknowledged or apparent from the method. \\
\bottomrule
\end{tabular}
\end{table}

\begin{table}[t]
\centering
\caption{Reviewer output schema (\texttt{scores.md}). The reviewer appends one verdict block per reviewed candidate.}
\label{tab:schema-reviewer}
\begin{tabular}{p{0.26\linewidth}p{0.66\linewidth}}
\toprule
Entry & Definition and instruction \\
\midrule
Verdict & One of survive, revise with the single weakest dimension to focus on, or reject with a short reason. \\
Evidence & A bullet list in which each bullet cites a paper id with a one or two sentence reason, or challenges a specific source or prior-work mechanism. \\
Strongest objection & The single biggest problem, stated in one sentence. \\
\bottomrule
\end{tabular}
\end{table}

\begin{table}[t]
\centering
\caption{Paper reader output schema (\texttt{papers/<id>.md}), a cumulative digest returned to the requesting agent.}
\label{tab:schema-reader}
\begin{tabular}{p{0.26\linewidth}p{0.66\linewidth}}
\toprule
Entry & Definition and instruction \\
\midrule
Title heading & The paper title. \\
What this paper does & One short paragraph of at most 120 words. \\
Goal-by-goal answers & One block per reading goal, each restating the goal and giving a focused answer of at most 300 words, grounding specifics in the relevant section or equation of the full text. \\
\bottomrule
\end{tabular}
\end{table}

\section{Training Details}

\subsection{Role-Specific Training and Rewards}
\label{sec:appendix-training}

\paragraph{Why we train roles separately.}
Our initial design optimized the complete pipeline using only the final proposal reward. The base model rarely completed the pipeline, so most examples received no informative signal. Training each role separately avoids this cold start and connects every update to a clearly defined artifact reconstructed from a human paper.

\paragraph{Role environments.}
Each training example contains the information expected at the corresponding stage of the harness. The gap finder receives a problem definition and challenge axis, the innovator receives a verified gap analysis, and the report writer receives an accepted candidate. Every role can search the offline database, request targeted readings, and revise its artifact over multiple turns using the same interface available at deployment.

\paragraph{Validity checks.}
Rubric rewards are computed only after an output passes the applicable structural and citation checks. For the gap finder and innovator, the rubric reward also requires full credit on a judged methodological-focus or methodological-contribution criterion. Gap analyses must remain on the assigned axis and contain no more than two gaps. Reports must complete every required field. Every cited paper must have been read during the current stage or supplied through a verified earlier stage.

\paragraph{Role-specific quality rubrics.}
The reference-free score captures methodological specificity, importance, feasibility, evidence, and clarity. The reference-grounded score compares the substantive fields of an output with the privileged reference while allowing a valid alternative to score well. The citation F1 score measures overlap between generated and reference citation sets. For every role, the complete reward is
\[
r=0.85\,g\left(\frac{r_{\mathrm{free}}+r_{\mathrm{ref}}+r_{\mathrm{cit}}}{3}\right)+0.15\,r_{\mathrm{proc}},
\]
where $g\in\{0,1\}$ is the validity gate and $r_{\mathrm{proc}}$ is a rule-based compliance term that penalizes protocol violations, including disallowed tool calls, invalid formats, and missing required fields. It does not positively reward the volume of searching or reading. The gap finder's reference-grounded score averages its fields uniformly; the innovator and report writer give important fields weight $2$ and optional fields weight $1$ (Table~\ref{tab:appendix-reward}).

\begin{table}[t]
\centering
\caption{Per-role validity gates. For every role, $r=0.85\,g(r_{\mathrm{free}}+r_{\mathrm{ref}}+r_{\mathrm{cit}})/3+0.15\,r_{\mathrm{proc}}$, where $g\in\{0,1\}$ and $r_{\mathrm{proc}}$ is the rule-based protocol-compliance term defined in Appendix~\ref{sec:appendix-training}. Thus a failed validity gate removes the rubric reward while preserving the compliance term. Reference-grounded fields are averaged uniformly for the gap finder and by importance tier for the innovator and report writer.}
\label{tab:appendix-reward}
\begin{tabular}{lp{0.72\textwidth}}
\toprule
Role & Validity gates (all must pass) \\
\midrule
Gap finder & Valid schema, assigned-axis match, at most two gaps, every cited ID resolves and was read, judged methodological focus \\
Innovator & Valid schema, one candidate that builds on the given gap, every cited ID resolves and was read, judged methodological contribution \\
Report writer & All required proposal fields present, every cited ID resolves and was read \\
\bottomrule
\end{tabular}
\end{table}

\paragraph{Judge and optimization summary.}
Qwen3.6-27B scores the rubric with reasoning disabled. Invalid judge responses are retried and excluded if they remain unparseable. A failed validity gate sets the rubric component to zero, while the small process-shaping term remains available. We train one LoRA adapter per role using asynchronous GRPO with a maximum policy staleness of one step and an importance sampling correction for the resulting staleness. Sequence lengths and the number of training steps are calibrated per role, and the full optimization configuration is reported in Appendix~\ref{sec:appendix-setup}.

\subsection{Training Design Choices}
\label{sec:appendix-reward-selection}
Before settling on the final reward and environment, we explored several alternatives.

\paragraph{Full-text reading over metadata.}
We compared giving the reader the paper full text against a faster metadata-only digest. The metadata variant collapsed training, with a near-zero mean reward and most rollouts truncated at initialization, against a mean reward around $0.5$ for the full-text reader, so we kept full-text reading and recovered throughput with asynchronous GRPO rather than by shortening the input.

\paragraph{Comparative innovator reward.}
We also tested a pairwise innovator reward that asks whether each candidate field is better than its reference counterpart, in place of the absolute reference-grounded score. This variant increasingly favored longer and more elaborate answers: its training reward spiked to ${\sim}0.83$, suggesting that the judge rewarded length and elaboration independently of substance, and then repeatedly collapsed. By contrast, the absolute-rubric reward remained stable around $0.66$, so we kept the absolute reference-anchored rubric.

\paragraph{Process shaping and validity gates.}
We keep a small rule-based compliance term that deducts reward for protocol violations such as disallowed tool calls, invalid formats, and missing required fields. We also experimented with richer per-turn process rewards, including a deterministic reward for retrieval overlap with the reference citations and the injection of a privileged-reference rollout into each group \citep{liu2025toward}. These were run as isolated experiments and not adopted into the final reward. We refined the validity gates over time. Gap depth moved from a soft penalty to a hard cap of two gaps per axis. The report writer switched from a reference list with anchors to bare inline citation IDs after about half of its early episodes failed the read-before-cite check. A brittle rule that rejected any numeric string was replaced by a judged penalty for stating predictions as measured results.

\paragraph{Optimization choices.}
We tuned optimization for stability under asynchronous GRPO, using an asymmetric probability ratio clip with no KL penalty and generation budgets calibrated to the observed rollout-length distribution. During development we also tried higher learning rates and smaller global batch sizes and varied the number of training epochs per role. The final values are reported in Appendix~\ref{sec:appendix-setup}.

\section{Evaluation Metrics and Judge Prompts}
\label{sec:appendix-evaluation-prompts}

\paragraph{Novelty measures.}
We use the three novelty names reported in Table~\ref{tab:endtoend}. \emph{All} is a 1--10 judgment of how much the proposal adds beyond the supplied closest papers from the complete comparison corpus, including pre- and post-cutoff work. \emph{Cutoff} applies the same judgment using only papers published before the target cutoff. \emph{In-domain} is the percentage of proposals that exhibit structural cross-setting transfer. A proposal receives one when it imports a mechanism from a different problem setting and adapts it through a shared underlying structure, and zero when the transfer is absent, superficial, or unsupported. Thus, unlike All and Cutoff, In-domain is a binary transfer judgment rather than a 1--10 comparison against a separate literature set. The innovator retrieval ablation additionally reports \emph{Mechanism non-obviousness}, which asks how readily the central mechanism follows from the three nearest pre-cutoff mechanism-side papers and standard techniques available at the cutoff. We implement this as two separate judge calls. The first receives the target problem, challenge, and prior papers but never the proposal, and independently derives candidate mechanisms. The second receives these frozen candidates and the proposal, then measures whether any ex-ante candidate reaches the proposal's central mechanism without adding a new hindsight-derived recipe. Four scored levels map to $\{0,1/3,2/3,1\}$ for trivial combination, routine extension, non-obvious step, and surprising, respectively; an invalid candidate set is unscored and retried. In-domain novelty and Mechanism non-obviousness are averaged over the fixed test-set denominator, with an unscored proposal contributing zero.

\paragraph{Proposal-quality measures and judge models.}
Relevance measures how directly the solution addresses the supplied problem and challenge; clarity measures organization and unambiguous presentation; specificity measures concrete technical detail; actionability measures whether the method and evaluation can be implemented \citep{liu-etal-2026-papermentor}; soundness measures the plausibility and internal consistency of the technical reasoning; and impact measures the importance of the contribution if successful. All are scored from 1 to 10. Clarity, specificity, actionability, soundness, and impact are assessed jointly from the proposal alone, whereas relevance is assessed with the target problem definition and challenge. The primary Qwen3.6-27B judge and the four corroborating judges---Claude-4.8, Gemma-4-31B, GPT-OSS-120B, and GPT-5.4---receive the same prompts and inputs. We therefore reproduce each distinct prompt once rather than duplicating it by judge model. Dynamic proposal, literature, and few-shot-example blocks are shown as angle-bracketed placeholders; line breaks are normalized for print.

\paragraph{Proposal quality: clarity, specificity, actionability, soundness, and impact.}
\begin{lstlisting}[style=judgeprompt]
SYSTEM:
You are a meticulous senior reviewer for a top-tier ML/NLP venue. You will be given the full text of a machine-generated research proposal / report. Judge it ON ITS OWN TERMS -- you have no reference paper or independent experimental results. Score each of the five dimensions below on an integer 1-10 scale and give a single concise sentence of reasoning for each. Judge only what is stated; do not invent supporting details or evidence.

Use these shared anchors: 1-2 = seriously deficient; 3-4 = major gaps; 5-6 = partly adequate with consequential gaps; 7-8 = strong with limited gaps; 9-10 = exceptionally strong and well supported by the supplied text.

Dimensions:
- actionability: Could a competent researcher actually execute this? Are the method steps, design choices, and experiments concrete enough to implement, not just aspirational?
- specificity: Is the proposal precise and detailed (concrete mechanisms, equations, named datasets/baselines/metrics) rather than vague or generic?
- clarity: Is it well-organized, unambiguous, and easy to follow? Are claims, notation, and contributions stated cleanly?
- impact: If it works, how significant is the contribution to the field? Does it address a problem that matters?
- soundness: Is the technical reasoning correct and internally consistent? Are claims supported, assumptions explicit, and failure modes acknowledged (no obvious errors or over-claiming)?

Return ONLY a JSON object of this exact shape:
{
  "actionability": {"score": <1-10>, "reasoning": "<one sentence>"},
  "specificity":   {"score": <1-10>, "reasoning": "<one sentence>"},
  "clarity":       {"score": <1-10>, "reasoning": "<one sentence>"},
  "impact":        {"score": <1-10>, "reasoning": "<one sentence>"},
  "soundness":     {"score": <1-10>, "reasoning": "<one sentence>"}
}

FEW-SHOT EXAMPLE:
<few-shot example>

USER:
# RESEARCH REPORT TO EVALUATE
<full generated proposal>
\end{lstlisting}

\paragraph{Proposal quality: relevance.}
\begin{lstlisting}[style=judgeprompt]
SYSTEM:
You are an expert reviewer assessing the RELEVANCE of a machine-generated proposed solution to the research problem it was asked to address. You are given the target PROBLEM DEFINITION and CHALLENGE, then the proposal (report).

Judge how directly and completely the proposed solution addresses THIS specific problem definition and challenge: a focused solution that squarely targets the stated challenge scores high; a solution that drifts to a different problem, only partially addresses it, or is generic scores low. Score relevance independently of novelty, correctness, and writing quality.

Use these anchors: 1-2 = unrelated or almost unrelated; 3-4 = weak or generic connection; 5-6 = addresses part of the challenge; 7-8 = directly addresses most of it; 9-10 = directly and fully targets it.

Return ONLY a JSON object:
{"relevance": {"score": <1-10>, "reasoning": "<one sentence on how well the solution targets the stated problem/challenge>"}}

FEW-SHOT EXAMPLE:
<few-shot example>

USER:
# TARGET PROBLEM DEFINITION
<problem definition>
# TARGET CHALLENGE
<challenge>
# PROPOSED SOLUTION (report)
<full generated proposal>
\end{lstlisting}

\paragraph{Novelty: All and Cutoff.}
\begin{lstlisting}[style=judgeprompt]
SYSTEM:
You are an expert reviewer assessing the NOVELTY of a machine-generated research proposal relative to the supplied closest related work. You are given the proposal and, as reference, the problem-definition, challenge, and solution of the existing papers that tackle the most similar problem. The comparison set may include papers published after the proposal's target cutoff; judge against the supplied references.

Judge how much the proposal contributes BEYOND this related work: genuinely new mechanisms/ideas/combinations score high; rediscovering, renaming, or trivially recombining what the reference papers already do scores low. Reward novelty that is real and non-obvious, not novelty-by-jargon.

Use these anchors: 1-2 = essentially the same central mechanism as a reference; 3-4 = minor adaptation or straightforward combination; 5-6 = a distinct step with substantial overlap; 7-8 = a materially different contribution; 9-10 = strongly distinct relative to the supplied references.

Return ONLY a JSON object:
{
  "novelty": {"score": <1-10>, "reasoning": "<one sentence naming what is new vs. what already exists in the references>"},
  "overlapping_prior_work": ["<short titles or ideas the proposal overlaps with>"]
}

FEW-SHOT EXAMPLE:
<few-shot example>

USER:
# PROPOSAL TO EVALUATE
<full generated proposal>
# CLOSEST PRIOR WORK (reference papers with similar challenge + problem definition)
<reference-paper blocks from the selected temporal comparison set>
\end{lstlisting}

\paragraph{Mechanism non-obviousness.}
\textit{Call 1: proposal-blind candidate derivation.}
\begin{lstlisting}[style=judgeprompt]
SYSTEM:
You are a senior researcher attempting to solve a target research problem using only the supplied prior papers and standard techniques available at the target cutoff. You have not seen, and must not speculate about, any proposal that will later be evaluated.

Independently derive up to three concrete candidate mechanisms that address the target challenge. Each candidate must specify:
- its central technical operation;
- a step-by-step derivation from the supplied papers and standard techniques;
- which reference supplies each nonstandard step; and
- the key assumption needed for the mechanism to work.

Do not merely restate the problem, list broad research directions, or use mechanisms unsupported by the supplied material. A new combination is allowed only when you explain concretely how its pieces compose. If no concrete mechanism can be derived, return an empty candidate list.

Return ONLY this JSON object:
{
  "candidates": [
    {
      "candidate_id": "C1",
      "central_mechanism": "<one sentence>",
      "derivation_steps": ["<step 1>", "<step 2>", "..."],
      "reference_uses": ["<paper title or id>: <piece supplied>"],
      "key_assumption": "<one sentence>"
    }
  ]
}

USER:
# TARGET PROBLEM DEFINITION
<problem definition>
# TARGET CHALLENGE
<challenge>
# NEAREST PRIOR WORK (all published before the target)
<three nearest mechanism-side prior-paper blocks>
Derive candidate mechanisms using only the information above. JSON only.
\end{lstlisting}

\textit{Call 2: comparison against frozen candidates.}
\begin{lstlisting}[style=judgeprompt]
SYSTEM:
You are a skeptical senior reviewer assessing whether a proposal's central mechanism was independently derivable before the proposal was revealed. You receive candidate mechanisms produced in a separate proposal-blind call, along with the source prior papers and the proposal under review.

The candidate list is frozen. Do NOT invent, repair, merge, or extend candidates after reading the proposal. Prior papers may be consulted only to verify a frozen candidate's stated derivation, not to construct a new proposal-conditioned recipe.

Work in this order.

STEP 1 -- State the proposal's central technical operation in one sentence.

STEP 2 -- Compare that operation with every frozen candidate. Select the closest candidate, if any, and identify the exact shared steps and the consequential steps absent from the candidate. If no candidate reaches the proposal mechanism, use NONE. Superficial topical similarity does not count as a derivation.

STEP 3 -- If the candidate list is empty, malformed, or too vague to support a step-level comparison, set candidate_set_valid to false and return unscored; do not infer non-obviousness from a failed derivation call. Otherwise, assign one scored level:
- "trivial_combination": a frozen candidate already contains the central mechanism through direct reuse or combination with no meaningful adaptation.
- "routine_extension": a frozen candidate reaches the central mechanism using only ordinary implementation steps, without a new structural observation.
- "non_obvious_step": every frozen candidate misses a consequential step or structural observation supplied by the proposal.
- "surprising": the missing consequential step also runs against a clear expectation established by the supplied papers.

Use the upper levels only when you can identify the consequential missing step. Do not reward impressive vocabulary, length, or confident phrasing.

Return ONLY this JSON object:
{
  "candidate_set_valid": true|false,
  "central_mechanism": "<one sentence>",
  "closest_candidate_id": "<candidate id, or NONE>",
  "shared_steps": ["<step present in both candidate and proposal>"],
  "missing_consequential_steps": ["<step absent from the candidate>"],
  "obviousness": "trivial_combination"|"routine_extension"|"non_obvious_step"|"surprising"|"unscored",
  "hardest_to_guess_step": "<one sentence, or none>",
  "confidence": <1-5>
}

FEW-SHOT EXAMPLE:
<few-shot example>

USER:
# PROPOSAL-BLIND CANDIDATES (verbatim output of Call 1)
<frozen candidate JSON>
# NEAREST PRIOR WORK (all published before the target)
<three nearest mechanism-side prior-paper blocks>
# PROPOSAL UNDER REVIEW
<proposal mechanism block>
Compare the proposal only against the frozen candidates, then grade it. JSON only.
\end{lstlisting}

\paragraph{Novelty: In-domain (structural cross-setting transfer).}
\begin{lstlisting}[style=judgeprompt]
SYSTEM:
You are a senior machine learning researcher judging whether a research proposal exhibits STRUCTURAL CROSS-SETTING TRANSFER. This binary judgment is reported as In-domain novelty. Do NOT judge the proposal's overall quality or correctness.

Assign score 1 only when ALL of the following are supported by the proposal:
1. It identifies a source problem setting that differs from the target setting.
2. It imports a concrete mechanism, abstraction, or analysis technique from that source setting. A citation or a general-purpose tool alone does not count.
3. It identifies a shared underlying structure that explains why the mechanism should transfer.
4. It specifies how the mechanism is adapted to exploit that structure in the target setting.

Otherwise assign score 0. This includes no cross-setting transfer, unsupported transfer claims, and superficial analogies that name another setting without establishing the shared structure and adaptation.

Support the judgment with one verbatim quote of at most 35 words from the proposal. If no quote supports all four requirements, assign score 0.

Return ONLY this JSON object:
{
  "source_setting": "<short phrase, or null>",
  "target_setting": "<short phrase, or null>",
  "transferred_mechanism": "<short phrase, or null>",
  "shared_structure": "<short phrase, or null>",
  "adaptation": "<short phrase, or null>",
  "evidence_quote": "<verbatim proposal span, <=35 words, or null>",
  "reasoning": "<one sentence>",
  "score": <0 or 1>
}

FEW-SHOT EXAMPLE:
<few-shot example>

USER:
# PROPOSAL
<proposal mechanism block>
Judge structural cross-setting transfer. JSON only.
\end{lstlisting}

\section{Data Construction and Ideation Process Decomposition}
\label{sec:appendix-label-generation}
Qwen3.6-27B generates the primary reference artifacts for all three stages. A small residual set of target papers, on which Qwen3.6-27B fails validation after regeneration, is completed with GPT-5.4. Table~\ref{tab:label-generation} reports the per-stage counts across the training, validation, and test partitions. GPT-5.4 accounts for only $76$ of the $44{,}931$ stage-level artifacts, under $0.2\%$ of the total, so the artifacts are overwhelmingly produced by the open model.

We fully deduplicate the partitions, ensuring that neither the validation nor test set overlaps the training set and that the validation and test sets are mutually disjoint.

\paragraph{Problem-formulation leakage audit.}
Because the problem definition and challenge are reconstructed from the complete target paper, theoretically they could inadvertently retain method-specific clues to the target solution if the LLM that performs this task doesn't follow instructions well. We manually inspected the extracted problem definition and challenge for 50 randomly sampled target examples, comparing each with its source paper. We found no instances ($0/50$) in which the input exposed the target method, including through method-specific terminology, assumptions, examples, or problem decompositions that would reveal the solution.

\begin{table}[t]
\centering
\caption{Number of accepted reference artifacts generated by each model, per stage, across the training, validation, and test partitions. Qwen3.6-27B produces the primary set, and GPT-5.4 completes a small residual set of target papers on which Qwen3.6-27B does not converge after regeneration.}
\label{tab:label-generation}
\begin{tabular}{lrrr}
\toprule
Stage & Qwen3.6-27B & GPT-5.4 & Total \\
\midrule
Gap finder & 14{,}970 & 7 & 14{,}977 \\
Innovator & 14{,}934 & 43 & 14{,}977 \\
Report writer & 14{,}951 & 26 & 14{,}977 \\
\midrule
All stages & 44{,}855 & 76 & 44{,}931 \\
\bottomrule
\end{tabular}
\end{table}

\section{Experiment Configuration}

\subsection{Training and Inference Configuration}
\label{sec:appendix-setup}
We train one LoRA adapter per role on the Qwen3.6-27B backbone with asynchronous GRPO. Each adapter uses rank $32$, scaling $\alpha=64$, and no dropout, applied to all linear modules except the output projection and the visual tower, which is unused for this text-only task. Optimization uses AdamW with a constant learning rate of $5\times10^{-6}$, weight decay $0.01$, no warmup, and gradient norm clipping at $1.0$, in bfloat16 with tensor parallel size $8$ on H100 GPUs. Each GRPO step draws $64$ prompts with $8$ generations per prompt, giving $512$ rollouts per step, and applies two gradient updates per step at a training batch of $256$ rollouts. Training uses an asymmetric probability ratio clip of $[0.2, 0.28]$ with no KL penalty in the reward, a maximum policy staleness of one step, and an importance sampling correction. Sequence lengths and training limits are calibrated per role: the gap finder uses a maximum sequence length of $26{,}624$ tokens for up to $2$ epochs or $800$ steps, the innovator $30{,}720$ tokens for up to $1$ epoch or $600$ steps, and the report writer $49{,}152$ tokens for up to $2$ epochs or $600$ steps. For each sub-agent, we select the checkpoint with the highest held-out validation reward; the test split is used only for final evaluation. The reported system uses gap-finder step 380, innovator step 310, and report-writer step 194. The judge is Qwen3.6-27B with reasoning disabled, sampled at temperature $0.7$ and top-$p$ $0.8$. At evaluation the orchestrator, reviewer, and paper reader use the base backbone while the three selected adapters are composed into \sysname{}-Trained. The trained variant and \sysname{}-Base share the same offline retrieval interface, output schema, and reading and generation budget.

\paragraph{Reading and generation budget (\sysname{} variants and backbone controls).}
Within the harness each producer sub-agent runs under a fixed per-turn generation budget and a per-role turn cap: the gap finder up to $15$ turns at $8{,}192$ tokens per turn, the innovator up to $20$ turns at $6{,}144$ tokens per turn, and the report writer up to $20$ turns at $16{,}384$ tokens per turn. The orchestrator is bounded by at most $100$ turns overall. Reading is done only by the paper reader, which runs as a single-turn tool with a $30{,}000$-token input cap per read and returns a digest. The three \sysname{}-Base controls (Qwen3.6-27B, Gemma-4-31B, and GPT-OSS-120B) run under these identical caps. The open source autoresearch systems and the general purpose agents run their own native agent loops, and we do not impose these caps on them. Each role is trained on $40$ nodes of eight H100 GPUs, with $32$ nodes dedicated to policy optimization and $8$ nodes to asynchronous rollout generation.

\subsection{Temporal Cutoff and Model-Memory Contamination}
\label{sec:appendix-temporal-contamination}
Our temporal split restricts retrieval to papers published before January 1, 2026, although knowledge stored in model parameters cannot be fully controlled. We randomly sampled 10 post-cutoff papers and queried each evaluated model about their contents, finding no evidence of reliable recall sufficient to compromise the evaluation; moreover, the task requires substantial reasoning and synthesis beyond memorizing a target paper. We therefore consider January 1, 2026 a reasonable cutoff that balances model capability with contamination risk.

\begin{table}[t]
\centering
\caption{Public release dates and sources for the models used in our experiments.}
\label{tab:model-release-dates}
\begin{tabular}{p{0.32\linewidth}p{0.27\linewidth}p{0.27\linewidth}}
\toprule
Model & Public release date & Release source \\
\midrule
Qwen3.6-27B & April 22, 2026 & GitHub release \\
Gemma-4-31B & April 2, 2026 & Google blog \\
GPT-OSS-120B & August 5, 2025 & OpenAI \\
GPT-5.4 & March 5, 2026 & OpenAI \\
Claude-4.8-Opus & May 28, 2026 & Anthropic \\
\bottomrule
\end{tabular}
\end{table}

As a direct screening check, we manually queried each of the five models about 10 randomly sampled post-cutoff target papers (50 model--paper queries in total). Each query supplied only the paper title and asked the model to describe the paper's contents, without retrieval or an abstract. None of the models identified or described any of the queried papers as known work. We use a title-only probe because supplying the abstract would itself reveal substantive content. This result provides evidence against obvious title-level memorization in the sampled cases, but it cannot detect latent, paraphrased, or method-level memory and therefore cannot rule out contamination.

We chose January 1, 2026 as a pragmatic tradeoff between model capability and temporal validity: it supplies a sufficiently broad pre-cutoff literature for this demanding task while reserving papers on or after that date from retrieval as future directions. The training, validation, and test partitions are all drawn from these post-cutoff targets, and the test partition remains held out from optimization. Accordingly, our results should be interpreted as performance under a controlled retrieval cutoff, not as proof that every evaluated model was parametrically unaware of every target paper. Residual model-memory contamination is a limitation of the retrospective evaluation.

\section{Additional Experimental Results}

\subsection{Role-Level Validation Reward}
\label{sec:appendix-rolelevel}
Table~\ref{tab:persubagent} compares each role's held-out validation reward for \sysname{}-Base with the reward at the checkpoint selected for \sysname{}-Trained. All three roles improve: the gap finder by $0.101$, the innovator by $0.107$, and the report writer by $0.105$. Figure~\ref{fig:reward_curves} shows the complete training and validation trajectories.

\begin{table}[t]
\centering
\caption{Held-out validation reward for \sysname{}-Base and at each role's selected \sysname{}-Trained checkpoint.}
\label{tab:persubagent}
\begin{tabular}{lrrrr}
\toprule
Role & Selected step & Base & Trained & $\Delta$ \\
\midrule
Gap finder & 380 & 0.566 & 0.667 & $+0.101$ \\
Innovator & 310 & 0.588 & 0.695 & $+0.107$ \\
Report writer & 194 & 0.786 & 0.891 & $+0.105$ \\
\bottomrule
\end{tabular}
\end{table}

\subsection{Training Diagnostics}

\begin{figure}[t]
\centering
\includegraphics[width=\textwidth]{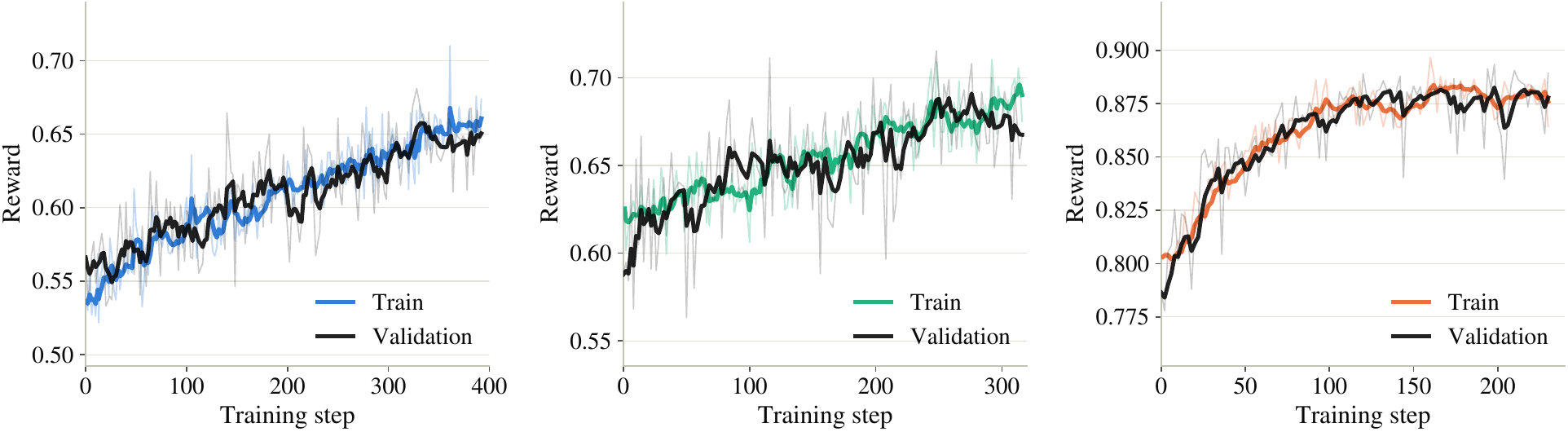}
\caption{GRPO reward over training for the gap finder (left, \colorsquare{gapcolor}), innovator (middle, \colorsquare{innovatorcolor}), and report writer (right, \colorsquare{writercolor}). Curves show the per-step training reward together with the held-out validation reward.}
\label{fig:reward_curves}
\end{figure}

\paragraph{Training and validation reward.}
Figure~\ref{fig:reward_curves} shows the per-step training and held-out validation rewards for all three trained roles.

\paragraph{Reward components.}
Decomposing the reward into its terms shows that the reference-free score begins high, while the reference-grounded and citation F1 scores provide more room for improvement.

\paragraph{Reward diversity.}
Across $27{,}225$ gap finder episodes, the median reward is $0.603$ and no reward is nonpositive. The mean within group standard deviation is $0.104$ across $3{,}473$ groups, and only $0.4\%$ of groups have zero variance. The reward therefore supplies a useful learning signal for nearly every sampled group.

\paragraph{Rollout length.}
At the selected gap-finder checkpoint (step 380), the $512$ training episodes have a median sequence length of $17.8$K tokens, a 95th percentile of $22.4$K, and a maximum of $25.8$K under the $26{,}624$-token limit. The median rollout contains $6$ turns and reads $6$ distinct papers; the corresponding 95th percentiles are $9$ turns and $9$ papers. Reward remains essentially uncorrelated with sequence length ($r=-0.024$), turns ($r=-0.031$), and papers read ($r=-0.017$).

\subsection{Generation Failure Analysis}
\label{sec:appendix-failures}
We recompute generation failures from the final 277-problem evaluation, defining a failure as a test problem without a scorable proposal. \sysname{}-Trained and all three open-backbone \sysname{}-Base runs have no failures, whereas Claude Code SDK has $5$ and Codex SDK has $8$.

\subsection{Checkpoint Selection}
\label{sec:appendix-checkpoint}
We trained each role over multiple checkpoints and evaluated more than one checkpoint combination. Table~\ref{tab:checkpoint-compare} compares the selected combination (gap finder step 380, innovator step 310, and report writer step 194) with the superseded combination (steps 122, 116, and 80) on the full test split under the same judge and metrics. The selected checkpoints improve valid generation, Citation F1, proposal quality, the reference-grounded score, the validity-gate pass rate, and the checkpoint-selection average.

\begin{table}[t]
\centering
\caption{Reported \sysname{}-Trained checkpoint versus an earlier checkpoint, scored by the same primary Qwen3.6-27B judge as the main results. Selection avg. averages all checkpoint-selection fields, including the proposal-quality dimensions, novelty scores, and field-level reference-grounded scores; it is distinct from Overall in Table~\ref{tab:endtoend}.}
\label{tab:checkpoint-compare}
\setlength{\tabcolsep}{3pt}
\begin{tabular}{@{}lrrrrrr@{}}
\toprule
Checkpoint & \shortstack{Valid\\gen.} & \shortstack{Citation\\F1} & Proposal & \shortstack{Ref.-\\grounded} & \shortstack{Gate\\pass} & \shortstack{Selection\\avg.} \\
\midrule
Earlier checkpoint & 0.993 & 0.037 & 0.791 & 0.259 & 0.993 & 0.531 \\
\textbf{\sysname{}-Trained (reported)} & \textbf{1.000} & \textbf{0.045} & \textbf{0.804} & \textbf{0.299} & \textbf{1.000} & \textbf{0.565} \\
\bottomrule
\end{tabular}
\end{table}

\section{Resource Use and Test-Time Scaling}

\subsection{Test-Time Compute}
\label{sec:appendix-test-time}
\begin{figure}[t]
\centering
\includegraphics[width=\linewidth]{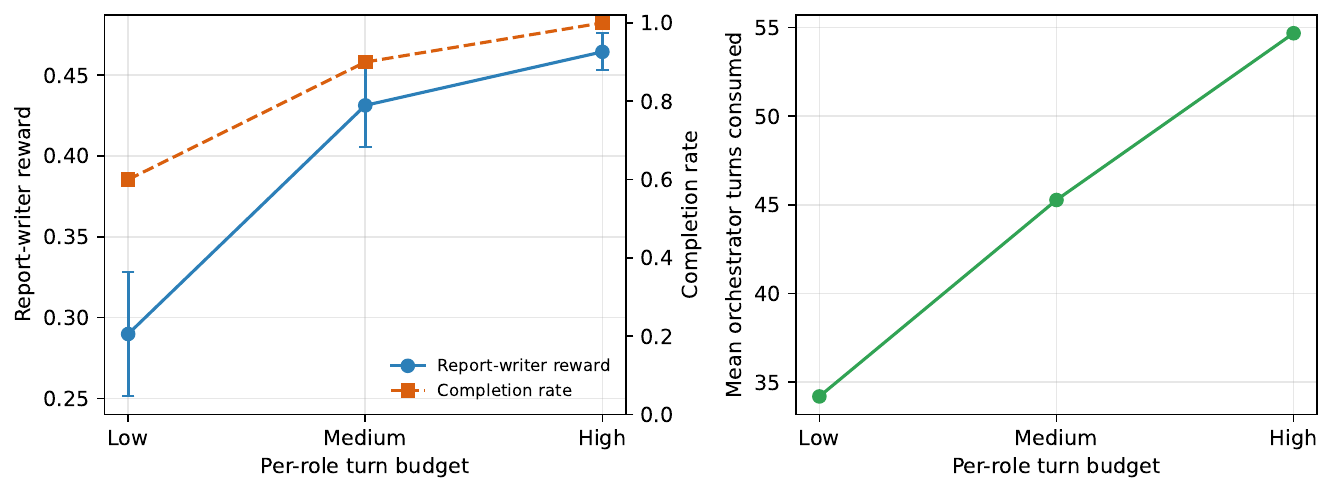}
\caption{Controlled test-time compute sweep on a fixed $40$ paper subset with \sysname{}-Base on Qwen3.6-27B. Only the per-role turn budget varies across the three levels. The reader settings, generation length, and retrieval are held constant. Left: mean report-writer reward with standard error and completion rate, the fraction of papers that produced a proposal and passed the validity gate. Right: mean orchestrator turns consumed, confirming that the low level is genuinely compute-starved. Reward and completion rise with budget, showing that the harness benefits from additional test-time compute.}
\label{fig:test-time-scaling}
\end{figure}

On the same Qwen3.6-27B backbone and harness, \sysname{}-Trained uses $168{,}916$ output tokens per delivered proposal, $11.1\%$ fewer than the $189{,}942$ used by \sysname{}-Base, so its gains do not result from a larger generation budget. Within the fixed per-role turn and token caps, the per-run generation budget still varies across papers. That variation is endogenous: harder papers can drive more turns while scoring lower, whereas runs that fail the validity gate early spend less and score zero. Correlating the report-writer reward with per-run turns, output tokens, or LLM calls under the caps therefore shows no positive association; it is small and slightly negative for \sysname{}-Trained and near zero for \sysname{}-Base, reflecting paper difficulty rather than a compute effect. To isolate the effect of compute, we ran a controlled sweep on a fixed subset of 40 test papers with \sysname{}-Base on Qwen3.6-27B, varying only the per-role turn budget across three levels while holding the reader settings, generation length, and retrieval constant. Figure~\ref{fig:test-time-scaling} shows that raising the budget from the low to the high level lifts the report-writer reward from 0.290 to 0.465 and the completion rate, the fraction of papers that yield a proposal passing the validity gate, from 0.60 to 1.00, while the mean orchestrator turns actually consumed rise from 34 to 55, confirming that the smallest budget is genuinely compute-starved. Additional test-time compute thus raises quality once the caps themselves are relaxed, complementing the role-training gains reported in the main results.

\section{Judge Reliability}
\label{sec:appendix-judge}
The system-level agreement in Figure~\ref{fig:judge-agreement}(b) is computed by ranking the generation systems under each judge's mean score for each metric family and taking the Spearman correlation with the leave-one-out mean of the other four judges, and with the Claude-4.8 judge, averaged across the metric families. The excess self-preference bias of a judge is the amount by which it rates its own model family's generations above the other judges' mean for that system, minus the judge's average deviation from the other judges across all systems, so a uniformly lenient or harsh judge is not counted as biased. At the individual-report level the judges cluster similarly: the score-level Pearson agreement with the leave-one-out consensus is $0.71$ for Qwen3.6-27B, $0.77$ for Claude-4.8, $0.77$ for Gemma-4-31B, $0.67$ for GPT-OSS-120B, and $0.78$ for GPT-5.4, placing every judge in the same reliable band.

\section{Case Studies}
\label{sec:appendix-case-studies}
The four cases below compare the human method, \sysname{}-Base, \sysname{}-Trained, and seven external systems under the same problem definition, model setting, inputs, and desired output. For the first three methods, each table separates the core idea, mechanism, and distinguishing contribution or limitation; the seven baselines are grouped by method family and summarized individually. The accompanying analysis focuses on how directly each mechanism addresses the stated challenge and reports when the evaluation scores do not support a strict ranking.

\subsection{Composable Safety Alignment}
{
\begin{longtable}{@{}p{0.22\textwidth}p{0.72\textwidth}@{}}
\caption{Case study on composable safety alignment. Every method receives a user question $x$ and a set $S$ of safety rules enabled for that request and must produce an answer that follows those rules. The language model's existing weights remain frozen.}
\label{tab:case-study-composable-safety}\\
\toprule
Component & Description \\
\midrule
\endfirsthead

\multicolumn{2}{@{}l}{\emph{Table~\ref{tab:case-study-composable-safety} continued}}\\
\toprule
Component & Description \\
\midrule
\endhead

\bottomrule
\endfoot

\textbf{Problem definition}
& Let $f_\theta$ be a language model whose weights $\theta$ remain fixed, and let $C=\{c_1,\ldots,c_K\}$ be the available safety categories. For each request, the system receives a user question $x$ and a selected subset $S\subseteq C$ specifying which categories are active. It must produce a response $y$ that follows the rules in $S$, including refusing or adapting an answer when an active rule requires it. It should still answer permitted questions helpfully and avoid refusing solely because of a category that is inactive. Small control components may be trained using examples of questions, active rule sets, and appropriate responses; the underlying model weights remain unchanged. \\
\midrule

\textbf{Challenge}
& Safety rules vary by user and context, but a model trained for one fixed policy cannot easily switch between them. Written instructions may not reliably enforce several rules at once. The challenge is to make one frozen model apply the rules selected for each request while still answering permitted questions.
\\
\midrule

\textbf{Human paper}
& \textbf{Core idea --- MOSAIC.} Represent every safety category with a small set of learned control tokens and prepend the tokens for exactly the rules enabled on the current request. \\*
& \textbf{Mechanism.} Train only these soft tokens on question--rule-set--answer examples, including harmless questions and repeated questions under different rule subsets; the language model remains frozen. \\*
& \textbf{Distinction / limit.} Policy switching happens entirely at the input, making composition simple and modular: unlike Base it does not alter hidden states, and unlike Trained it performs no generation-time checking. Success still depends on the frozen model composing the tokens reliably. \\
\midrule

\textbf{\sysname{}-Base}
& \textbf{Core idea.} Model the enabled policies as a set whose members can interact, rather than adding independent control vectors: a Set Transformer produces an interaction-aware policy representation. \\*
& \textbf{Mechanism.} Train rule embeddings, the set aggregator, and a FiLM controller on matched rule-set responses; FiLM uses the aggregate to scale and shift selected hidden signals while the language model stays frozen. \\*
& \textbf{Distinction / limit.} Base is the only proposal here that explicitly learns nonlinear interactions among the active rules before steering. This may capture conflicts missed by additive methods, but it needs combination coverage and can destabilize pretrained representations \citep{liu2026mixsd}. \\
\midrule

\textbf{\sysname{}-Trained}
& \textbf{Core idea.} Treat composition as constraint satisfaction: learn an acceptable-state boundary for each rule and move the current activation into the intersection of all enabled boundaries. \\*
& \textbf{Mechanism.} Train per-rule probes from acceptable and violating generations; at decoding time, solve for a nearby activation satisfying every active probe, then continue generation from the corrected state. \\*
& \textbf{Distinction / limit.} Unlike weighted vectors, adapters, or reward scores, Trained seeks joint satisfaction rather than a compromise and minimizes activation displacement. Linear probe boundaries are only behavioral proxies, however, and feasible/nearest projection is not automatically guaranteed. \\
\midrule

\multicolumn{2}{@{}l}{\textbf{Baselines: modular adapters}} \\*
\textbf{AutoResearchClaw}
& Parses the active rules and routes through category-specific, disentangled adapters; a lightweight router combines the selected modules before frozen-model generation. \\

\textbf{DeepScientist}
& Trains one nearly orthogonal LoRA adapter per category and an input-aware network to weight active adapters, optionally projecting to preserve general capabilities. \\
\midrule

\multicolumn{2}{@{}l}{\textbf{Baselines: conditioning and additive steering}} \\*
\textbf{ResearchAgent}
& Encodes active rules as conditioning embeddings and guides the frozen model through a safety router, switching policy through the conditioning signal. \\

\textbf{CycleResearcher}
& Learns one logit-steering vector per category, combines active vectors with a policy encoder, and trains first on single rules and then combinations. \\

\textbf{AI-Researcher}
& Combines rules into a hidden-state control vector, removes components associated with general capabilities, and context-gates the intervention strength. \\
\midrule

\multicolumn{2}{@{}l}{\textbf{Baselines: decoding-time scoring or instructions}} \\*
\textbf{Dr.~Claw}
& Trains a separate autoregressive reward model per category and combines the enabled models' scores during decoding; a policy manager selects the active set. \\

\textbf{SciAgents}
& Classifies a safety action from the active policies and writes a refusal or redirection instruction, relying on the frozen model to follow that instruction. \\

\end{longtable}
}

The human paper composes safety policies with learned input tokens, whereas Base models interactions among the active rules with a Set Transformer and FiLM. Trained changes the formulation more fundamentally: each rule defines a learned boundary over model states, and generation projects toward their joint feasible region. The baselines instead compose adapters, additive steering signals, reward scores, or generated instructions; even AI-Researcher's projection protects a capability subspace rather than enforcing the intersection of active policy constraints.

\subsection{Adaptive Rollout Allocation}
{
\begin{longtable}{@{}p{0.22\textwidth}p{0.72\textwidth}@{}}
\caption{Case study on adaptive rollout allocation for reasoning RL. Every method uses training questions with reference answers or an automatic correctness checker together with a student model. The methods differ in how they spend the rollout budget and construct the groups used for GRPO.}
\label{tab:case-study-adaptive-rollout-allocation}\\
\toprule
Component & Description \\
\midrule
\endfirsthead

\multicolumn{2}{@{}l}{\emph{Table~\ref{tab:case-study-adaptive-rollout-allocation} continued}}\\
\toprule
Component & Description \\
\midrule
\endhead

\bottomrule
\endfoot

\textbf{Problem definition}
& Given a dataset $D$ of question--answer pairs $(q,a)$ and a student policy $f_\theta$, the model generates multiple answers, or \textbf{rollouts}, for each sampled question. A verifier compares each answer with the reference answer and assigns a binary correctness reward.

The goal is to improve the student's reasoning accuracy under a fixed compute budget by deciding which questions receive additional rollouts and how their samples contribute to training. GRPO calculates advantages by comparing rewards within each question's rollout group. The baseline objective also includes a KL penalty to limit deviation from a reference policy.

\textbf{Inputs.} A batch of questions with reference answers, the student policy, a compute budget, and the generated rollouts with their token log probabilities and correctness rewards.

\textbf{Output.} Updated student policy parameters, evaluated on the model's ability to answer questions independently.

Every method uses training questions with reference answers or an automatic correctness checker together with a student model. A \textbf{rollout} is one answer generated by the student, including its reasoning. Several rollouts for the same question form a \textbf{group}. GRPO compares their rewards to calculate an \textbf{advantage}, which is positive for answers that perform better than the group average and negative for those that perform worse. If every answer receives the same reward, their relative reward advantages are zero. \\
\midrule

\textbf{Challenge}
& Standard GRPO training typically assigns the same number of rollouts to every question, despite differences in difficulty and potential training value. This can waste computation on consistently solved questions while giving difficult but learnable questions too few attempts to produce a correct answer and a useful training signal. The challenge is to allocate a fixed sampling budget more effectively to improve reasoning performance. \\
\midrule

\textbf{Human paper}
& \textbf{Core idea --- CoDaPO.} Prioritize questions using both empirical difficulty (fraction incorrect) and token-probability confidence, favoring questions that are moderately difficult but that the student appears ready to learn. \\*
& \textbf{Mechanism.} Generate an initial rollout group for every question, repeat the highest-valued questions to form a fresh second batch, and use the same values to weight the training loss; no teacher is required. \\*
& \textbf{Distinction / limit.} CoDaPO changes both \emph{which questions are resampled} and \emph{how strongly they train the model}; unlike the methods below, it couples allocation with loss weighting. A fair efficiency comparison must charge both rollout stages. \\
\midrule

\textbf{\sysname{}-Base}
& \textbf{Core idea.} Treat questions as arms in a UCB-style allocator: empirical success estimates current solvability, while an uncertainty bonus gives under-sampled questions an optimistic chance to receive more compute. \\*
& \textbf{Mechanism.} Generate an initial group per question, repeatedly assign the remaining budget to the highest upper-confidence score, and train GRPO on the resulting variable-size whole-question groups; no teacher model is required. \\*
& \textbf{Distinction / limit.} The exploration bonus can sample apparently very hard questions that variance-based rules discard, then decays if evidence shows they are unreachable. The rule is standard bandit machinery and assumes question difficulty is sufficiently stationary as the policy changes. \\
\midrule

\textbf{\sysname{}-Trained}
& \textbf{Core idea.} Estimate distance to the learning frontier from partial reasoning: prefix-conditional success separates questions that are close to a solution from those that are merely hard and currently unreachable. \\*
& \textbf{Mechanism.} From a small initial group, take the most confident trajectory and an intermediate prefix, sample short continuations, and classify the question as saturated, learnable, or beyond reach; use that estimate to allocate fresh full rollouts. \\*
& \textbf{Distinction / limit.} Unlike baselines driven by final-answer statistics or gradients, Trained asks whether a promising partial solution can be completed. Its diagnostic continuations are discarded rather than trained on, and estimates from a small sample and fixed regime thresholds may be noisy. \\
\midrule

\multicolumn{2}{@{}l}{\textbf{Baselines: outcome-statistic schedulers}} \\*
\textbf{AutoResearchClaw}
& Uses recent success-rate variation as a learnability signal, while reserving some easy questions for calibration and behavioral stability. \\

\textbf{Dr.~Claw}
& Combines rollout accuracy, reward variance, and entropy to favor hard-but-learnable questions; it also relaxes KL regularization for hard questions and strengthens it for easy ones. \\

\textbf{ResearchAgent}
& Allocates from early reward variance and verifier feedback, then normalizes advantages within difficulty bands to stabilize unequal group sizes. \\

\textbf{CycleResearcher}
& Allocates from initial confidence and reward variance, and rescales advantages by the allocation ratio to balance gradients across variable-size groups. \\

\textbf{AI-Researcher}
& Uses moving averages of success and reward variance to estimate learning potential, with modified advantage normalization for unequal rollout counts. \\
\midrule

\multicolumn{2}{@{}l}{\textbf{Baselines: learned or gradient-based value signals}} \\*
\textbf{SciAgents}
& Trains a lightweight difficulty predictor to drive a capped rollout scheduler, retaining within-question advantage normalization for stability. \\

\textbf{DeepScientist}
& Values questions by the magnitude and cross-trajectory agreement of their policy gradients, then assigns both rollouts and loss weight accordingly; this requires per-question gradient statistics. \\

\end{longtable}
}

The human paper jointly resamples and loss-weights questions using confidence and difficulty. Base adds a standard UCB exploration bonus to whole-question allocation, while most baselines schedule from success, variance, entropy, or a learned difficulty signal; DeepScientist is the exception, using gradient magnitude and agreement. Trained alone measures whether a promising partial trajectory can be completed, which directly targets the hard-but-learnable versus unreachable distinction.

\subsection{Capability Grounded Subagent Abstention}
{
\begin{longtable}{@{}p{0.22\textwidth}p{0.72\textwidth}@{}}
\caption{Case study on capability grounded subagent abstention. Every method receives a user request and information about what the subagent can do. The methods differ in how they decide whether to answer or abstain. The same example applies throughout. A user asks for customer groups by age and gender, but the subagent cannot access those fields.}
\label{tab:case-study-subagent-abstention}\\
\toprule
Component & Description \\
\midrule
\endfirsthead

\multicolumn{2}{@{}l}{\emph{Table~\ref{tab:case-study-subagent-abstention} continued}}\\
\toprule
Component & Description \\
\midrule
\endhead

\bottomrule
\endfoot

\textbf{Problem definition}
& A coordinator sends a user request $q$ to a specialized sub-agent. The sub-agent receives the request, its task context, and information about its available tools, data, and permissions. It must return either a valid task result $y$, such as a customer segmentation query, or a structured abstention $a$. The abstention states the failure category and the specific reason the task cannot be completed, such as a missing age field, a privacy restriction, or an unclear request. The goal is to avoid unsupported answers while continuing to complete requests the sub-agent can handle. \\
\midrule

\textbf{Challenge}
& Specialized sub-agents sometimes give plausible answers to requests they cannot complete because the request is unclear or requires unavailable tools, data, or permissions. The coordinator may pass those answers to users without knowing they are unsupported. The challenge is to help sub-agents recognize these limits and report them clearly enough for the coordinator to clarify, reroute, or stop the request. \\
\midrule

\textbf{Human paper}
& \textbf{Core idea --- EARS.} Teach abstention as part of the sub-agent's ordinary response policy, covering missing capabilities, ambiguous requests, and requests sent to the wrong specialist. \\*
& \textbf{Mechanism.} Fine-tune on production requests paired with either valid results or typed abstentions; humans label a seed set and calibrated LLM judges scale the annotations. At inference, the sub-agent decides directly whether to answer. \\*
& \textbf{Distinction / limit.} Unlike registry gates, EARS can learn limits that are difficult to formalize and needs no separate checker on every request. Its decisions are not mechanically grounded in a live inventory and depend on training-data coverage. \\
\midrule

\textbf{\sysname{}-Base}
& \textbf{Core idea.} Verify the \emph{reason} for abstention, not only the answer/abstain decision, so the coordinator receives a minimal, causally relevant explanation of the failure. \\*
& \textbf{Mechanism.} Train on examples of checker feedback and corrected explanations; at inference, an LLM translates its proposed reason into inventory conditions and revises it until a logic checker confirms both necessity and relevance. \\*
& \textbf{Distinction / limit.} Baseline gates mostly stop at detecting infeasibility; Base adds iterative proof-checking of the stated reason. Its reliability still rests on an accurate inventory and faithful translation from language to logic. \\
\midrule

\textbf{\sysname{}-Trained}
& \textbf{Core idea.} Prevent unsupported generation by decomposing the request into required tools, fields, permissions, and operations, then checking every requirement before the sub-agent runs. \\*
& \textbf{Mechanism.} Compare the LLM-extracted requirements with a live structured inventory; execute only if all checks pass, otherwise return a typed abstention naming the missing or prohibited capability. The basic version needs no training. \\*
& \textbf{Distinction / limit.} This is an explicit, auditable gate rather than EARS's learned decision, but it is methodologically close to five baselines below. Its main value is the concrete requirement decomposition; errors in parsing or stale inventories remain failure points. \\
\midrule

\multicolumn{2}{@{}l}{\textbf{Baselines: explicit pre-execution capability gates}} \\*
\textbf{Dr.~Claw}
& Checks the request against a declared capability list and returns a typed failure reason that the coordinator can use to reroute, clarify, or stop. \\

\textbf{ResearchAgent}
& Matches request requirements to a registry of tools and constraints before execution, returning a structured failure category and explanation on mismatch. \\

\textbf{CycleResearcher}
& Places a thresholded feasibility assessor over a capability schema before the task agent, choosing execution or structured abstention. \\

\textbf{SciAgents}
& Verifies required tools, data, and permissions before execution and reports a categorized reason for coordinator rerouting or escalation. \\

\textbf{DeepScientist}
& Checks each request against explicit tools, constraints, and scope boundaries before generation, producing typed abstentions for downstream recovery. \\
\midrule

\multicolumn{2}{@{}l}{\textbf{Baselines: confidence-aware and hybrid gates}} \\*
\textbf{AutoResearchClaw}
& Maps a calibrated pre-execution feasibility score to risk, then also audits outputs so the coordinator can decompose, reroute, or escalate failures. \\

\textbf{AI-Researcher}
& Combines executable hard constraints with semantic capability matching and an uncertainty gate, returning recovery guidance when the check fails. \\

\end{longtable}
}

The human paper fine-tunes abstention as part of the sub-agent's response policy, allowing it to learn ambiguous or misrouted cases but tying reliability to data coverage. Base instead checks and revises the stated reason for failure, emphasizing a precise explanation. Trained decomposes a request into required capabilities and checks them against a live inventory before generation. Five baselines use essentially the same registry-gate pattern; AutoResearchClaw adds output auditing, and AI-Researcher combines executable constraints with semantic matching and uncertainty.

\subsection{Adaptive Multimodal Speculative Decoding}
{
\begin{longtable}{@{}p{0.22\textwidth}p{0.72\textwidth}@{}}
\caption{Case study on adaptive drafting for multimodal speculative decoding. Every method receives an image, a conversation, and the tokens already accepted by a target model. The methods differ in how they adapt drafting when a conversation shifts between visually grounded and text only requests.}
\label{tab:case-study-adaptive-speculative-decoding}\\
\toprule
Component & Description \\
\midrule
\endfirsthead

\multicolumn{2}{@{}l}{\emph{Table~\ref{tab:case-study-adaptive-speculative-decoding} continued}}\\
\toprule
Component & Description \\
\midrule
\endhead

\bottomrule
\endfoot

\textbf{Problem definition}
& A large vision and language model is the \textbf{target} whose answer we want to produce; a cheaper model is the \textbf{drafter} that proposes several tokens at a time. Given the image, the text prompt, and the tokens already accepted, a method chooses how to draft the next block of tokens. The target checks that block in one verification pass. Accepted tokens become part of the answer; when a token is rejected, the verification procedure supplies a corrected token. The objective is to reduce total generation time while preserving the target model's output distribution through the standard acceptance and correction procedure. The method produces draft probabilities as an intermediate result and an accepted token sequence as its final result \href{https://proceedings.mlr.press/v202/leviathan23a}{(Leviathan et al., 2023)}. \\
\midrule

\textbf{Challenge}
& A draft model that predicts well for image questions may predict poorly when the conversation shifts to a question that needs only text. Poor predictions make the large model reject more draft tokens, while extra drafting work can erase any speed gain. The challenge is to adapt drafting to the current request so that the \textbf{time to produce the final answer} improves across different kinds of conversation. \\
\midrule

\textbf{Human paper}
& \textbf{Core idea --- TABED.} Keep visual and text drafting active as an ensemble, and continuously shift their mixture weights toward whichever method has recently predicted the target model better. \\*
& \textbf{Mechanism.} Combine the drafters' next-token distributions; after each target verification, compare their saved predictions with the target distribution and update the weights for the next block. No new router training is required. \\*
& \textbf{Distinction / limit.} TABED uniquely learns online from \emph{past target verification} while retaining multiple drafters, rather than making a hard routing decision. This robustness costs extra drafting work even with shared parameters and batching. \\
\midrule

\textbf{\sysname{}-Base}
& \textbf{Core idea.} Use the divergence between a drafter's text-only and vision-conditioned next-token distributions as a step-level estimate of how much information the image contributes. \\*
& \textbf{Mechanism.} Compute both draft distributions, turn their KL divergence into a gate on visual-feature injection, and close the visual path when it is redundant while retaining standard target verification. No new training is required. \\*
& \textbf{Distinction / limit.} Base proposes continuous visual gating rather than selecting a fixed specialist, making it finer-grained than the routing baselines. Running both draft paths adds cost, and the written pseudocode does not clearly connect the gate to the final draft-token distribution. \\
\midrule

\textbf{\sysname{}-Trained}
& \textbf{Core idea.} Route each conversational turn using the target model's own prefill-stage language-to-visual attention mass as a zero-shot estimate of visual relevance. \\*
& \textbf{Mechanism.} Normalize the already-computed attention signal, choose either a vision-conditioned or text-only specialist for the turn, and pass that drafter's blocks through ordinary target verification; the router needs no training. \\*
& \textbf{Distinction / limit.} Unlike heuristic modality classifiers, Trained reads the model whose distribution the drafter must approximate; unlike TABED, it commits to one drafter rather than maintaining an ensemble. Attention is still only a proxy, and target prefill is not literally zero-latency overhead. \\
\midrule

\multicolumn{2}{@{}l}{\textbf{Baselines: hard routing among specialized drafters}} \\*
\textbf{AutoResearchClaw}
& Classifies request modality, selects a specialized drafter, prunes visual tokens, and falls back to ordinary decoding when draft confidence is low. \\

\textbf{Dr.~Claw}
& Uses a lightweight, training-free context classifier to switch between visual and text-reasoning drafters as the request type changes. \\

\textbf{SciAgents}
& Uses attention entropy and token diversity to label each phase visual or textual, then sends the chosen specialist's block to the target for verification. \\

\textbf{DeepScientist}
& Detects visual, mixed, and textual phases from token proportions and positions, switching specialist drafters only at block boundaries. \\
\midrule

\multicolumn{2}{@{}l}{\textbf{Baselines: multi-signal routing and adaptive control}} \\*
\textbf{ResearchAgent}
& Uses estimated modality balance and reasoning complexity to select or blend drafters, with visual-token pruning and a syntax-aware text drafter. \\

\textbf{CycleResearcher}
& Routes from attention entropy and key--value-cache gradients, while adapting draft length to maintain the target acceptance rate. \\
\midrule

\multicolumn{2}{@{}l}{\textbf{Baseline: online soft ensemble}} \\*
\textbf{AI-Researcher}
& Mixes visual, textual, and hybrid draft distributions, adjusts the visual-token budget, and updates mixture weights from recent verification outcomes without gradient training. \\

\end{longtable}
}

The human paper keeps several drafting methods active and updates their mixture weights from past target verification. Base instead computes text-only and vision-conditioned draft distributions at each step and uses their KL divergence to gate visual-feature injection; its pseudocode does not clearly connect that gate to the final draft-token distribution. Trained makes a hard per-turn choice from the target model's own prefill-stage visual attention, avoiding a learned or heuristic context classifier and the human method's continuing ensemble cost. The baselines mostly route from modality or phase indicators; ResearchAgent and CycleResearcher add richer control signals, while AI-Researcher is the closest online ensemble.

\end{document}